\documentclass[5p,times]{elsarticle}

\usepackage{amssymb}

\usepackage{graphicx}
\usepackage[utf8]{inputenc}
\usepackage[T1]{fontenc}
\usepackage{caption}
\usepackage{subcaption}
\usepackage{color}
\usepackage{verbatim}
\usepackage{hyperref}
\usepackage{todonotes}

 \usepackage{makecell}
\usepackage{multirow}
\usepackage{array}

 \usepackage{hhline}
 \usepackage{rotating}
 \usepackage{comment}

\newcommand{\Input}{{\hspace*{\algorithmicindent} \textbf{input }}}

\usepackage[nolist]{acronym}
\begin{acronym}[DRAM]
\acro{TA}{Tsetlin automaton}
 \acrodefplural{TA}{Tsetlin automata}
 \acro{TM}{Tsetlin machine}
    \acro{RTM}{Relational Tsetlin machine}
\end{acronym}

\usepackage{algorithm}
\usepackage{algpseudocode}

\journal{Data \& Knowledge Engineering}

\begin{document}

\begin{frontmatter}



\title{Efficient Data Fusion using the Tsetlin Machine}


\author[inst1]{Rupsa~Saha}

\affiliation[inst1]{organization={Centre for AI Research,Department of IKT},
            country={Norway}}
            
\author[inst2,inst1]{Vladimir~I.~Zadorozhny}
\author[inst1]{Ole-Christoffer~Granmo}

\affiliation[inst2]{organization={School of Computing and Information, University of Pittsburgh},
            country={USA}}

\begin{abstract}
We propose a novel way of assessing and fusing noisy dynamic data using a Tsetlin Machine. Our approach consists in monitoring how explanations in form of logical clauses that a TM learns changes with possible noise in dynamic data. This way TM can recognize the noise by lowering weights of previously learned clauses, or reflect it in the form of new clauses. We also perform a comprehensive experimental study using notably different datasets that demonstrated high performance of the proposed approach. 
\end{abstract}


\end{frontmatter}

\section{Introduction}

The amount of data available due to the rapid spread of advanced information technology is exploding.  It is not uncommon for digital archives to store petabytes of data in hundreds of data repositories supporting thousands of applications. The volume of data storage grows in excess of 50\% annually. This growth is expected to continue due to new Web-based systems, increased utilization of existing systems, and the introduction of new sources of data. This dynamic motivates continued research on Data Integration Systems to provide users with uniform data access and efficient data sharing. 

The ability to share data is particularly important for interdisciplinary research, where a comprehensive picture of the subject requires large amounts of data from disparate data sources from a variety of disciplines. For example, epidemiological data analysis often relies upon knowledge of population dynamics, climate change, migration of biological species, drug development, etc.  As another example, consider the task of exploring long-term and short-term social changes which requires consolidation of a comprehensive set of data on social-scientific, health, and environmental dynamics.
Meanwhile, the data sources may have different levels of reliability for many reasons, e.g., issues with the primary sources of information, faulty data collection methodology, etc. As a result, integration of the data sources may also face severe data inconsistencies. 

Reliable consolidation of diverse data sources requires applying advanced data fusion strategies for efficient resolution of data inconsistencies. In many cases, such inconsistency can be discovered through analysis of relationships between the data sources. Small quantities of such inconsistencies can be regarded as noise. However if the inconsistencies start becoming more and more prevailing, it is a good idea to regard them as part of the data itself, rather than as aberrations. Thus, we need to perform proper data fusion to collate data from different sources or backgrounds. These differences can result in data with differing characteristics that need to undergo fusion. While the data overall can pertain to the same theme or task, the differing characteristics can lead to challenges on further processes carried out on the data.

In this paper we propose a novel approach to fusion of the data sources with conflicting and inconsistent information based on the concept of Tsetlin Machine (TM). The TM implements advanced interpretable learning
that aims at discovering (learning) logical clauses explaining the data.
Our approach efficiently utilizes Tsetlin Machine for tracking differences and changes in data sources due to its capabilities in obtaining a logical description (clauses) of the data at hand. The clauses learnt as a whole (i.e. the global description) on two alternative data sources can give us an overview of the differences in the two datasets, which can be further used to fuse them. On the other hand, clauses pertaining to individual instances (i.e. the local description) can be useful in identifying data coming from sources that differ in some aspect. While the overall objective is to differentiate data based on their characteristics, recognizing the differences can be challenging due to the nature and source of the differences. Therefore we break down this task into three smaller sub-goals:

\begin{itemize}

\item Acknowledging differences and changes in data : \textit{can we say that current data is not the same as data handled previously?}
\item Identifying changed and conflicting data : \textit{can we localize which data samples are not the same as the previously handled data?}
\item Mitigating or amplifying changes in order to resolve conflicts and provide a most likely data interpretation :\textit{ can we process the identified data in some way which benefits the original goal of training with the data?}

\end{itemize}

The paper is organized as follows: In the next section (section \ref{sec:background}) we talk about the existing data fusion area of research, as well as about Tsetlin Machines, which are the cornerstone of our approach. In section \ref{sec:Proposal}, we outline our Tsetlin Machine based approach of recognizing and dealing with data that the model has not been exposed to previously. 
Subsequently, section \ref{sec:Real Experiments} provides an comprehensive experimental study on different datasets, viz. Network Intrusion data, 20 Newsgroup Data and patient health data to showcase the nuances of the proposal in varied real world scenarios. Finally, we summarize our findings in section \ref{sec:Conclusion}

\section{Background and Related Work}
\label{sec:background}
\subsection{Data fusion}
Data fusion, or more generally, information fusion (IF), has been an area of research and development since the 1980s, and have mainly been associated with the sensors subspace within machine learning. A broad definition of IF describes it as the process of combining data and information from different sources and assessing their significance. The fusion process is necessarily dynamic, and provides continuously updated estimates and assessments. The goal of fusion is to ultimately achieve better results than what could be obtained without performing the fusion \cite{snidaro2019recent}.

Data Integration systems that process information from more than one source generally require one or more of the following steps of processing techniques on the source data: (1) data transformation, (2) duplicate detection and (3) data fusion \cite{bleiholder2009data}. Data transformation is essential when the data from differing sources do not conform to one standard global schema. Duplicate detection is required to resolve multiple references to the same object via different sources, and if necessary, identify and resolve resultant ambiguities. Based on the results of the previous two steps, further resolving data conflicts due to data inconsistencies is related to data fusion.

Data transformation and duplicate detection can take care of conflict arising from the external structure of the data, i.e. issues arising from different representations of same entity by different sources. Data fusion handles data conflicts caused by various data inconsistencies. Data conflicts may take the form of semantically equivalent attributes from different sources having conflicting attribute values for same (or similar) entities. A simplistic example would be conflict while merging Source 1 and Source 2 for data describing characteristics of successful basketball players, and Source 1 reporting average height as 'greater than 175cm' and Source 2 reporting average height (i.e. same attribute) as '165 cm'. Data conflict is usually of two types : (1) uncertain attribute value due to missing information, and (2) contradicting attribute values.  

Strategies for handling data conflicts can be one of the following:
\begin{itemize}
    \item Conflict ignoring - As the name implies, this strategy is simply to be unaware of any conflicts. Decisions regarding resolution is often left to the discretion of the human user.
    \item Conflict avoidance - In contrast to the previous strategy avoidance involves, at the very least, acknowledgement of conflicting information. A prefabricated and universal decision mechanism usually takes over at that point, for example, deferring to known trustworthiness values of the conflicting sources.
    \item Conflict resolution - In this strategy, resolution is decided taking into account the conflicting information itself and the surrounding metadata. Some techniques under this strategy can further be classified as either deciding (choosing one over another) or mediating (choose a value representative of both options, but not necessarily part of either).
\end{itemize}

While the above procedure is followed for integrated databases, sensor applications is another area where data fusion techniques are heavily required, and the existing literature highlights methods that are somewhat different than those discussed above. According to \cite{castanedo2013review}, data fusion techniques in a multisensor scenario broadly fall in one of the following categories : (1) data association (2) state estimation and (3) decision fusion. These are briefly described as follows. 
Data association deals with trying to determine associations between the data, the reporting sensor and the target that purportedly generated the data. This becomes necessary when a single sensor tracks multiple targets, and it is important to differentiate the received data. State estimation methods are used to determine the state of the target producing the data, to determine is the data is valid (or relevant) based on the state. The decision fusion methods are the most high level techniques. They usually involve some sort of logic and external information in order to account for inconsistencies in the environment \cite{castanedo2013review}. 

At the most conceptual level, data fusion characteristics are captured by the JDL Data Fusion Classification and its subsequent remodels \cite{kessler1992functional, llinas2004revisiting, blasch2013revisiting}. It does not propose a particular technique, rather it aims to provide an overview with which data fusion architecture can be designed. 
As such, the JDL does not offer methods of data fusion, but perspectives from which the fusion task can be approached (and solved). The five levels ( 0 to 4) are: source preprocessing, object refinement, situation assessment, impact assessment and finally, process refinement. 

As discussed above, most of existing literature deals with data fusion with respect to sensors in a multisensor environment, or with relational databases. However, with explainable AI systems slowly coming to the fore, there is now a wider scope, namely, the integration of AI and data fusion. Data fusion can allow AI models to better integrate context in their decisions, making it possible to have a human understandable representation of the decision taken by the model. The explainable behaviour afforded by early AI (which was primarily expert crafted) can be conjoined with the power of modern learning algorithms via such explainable models. For example, previously expert formed keywords related to causality and sentiment analysis was shown to be validated as well as enhanced by the use of TM-based learning mechanism \cite{saha2020mining}. Parallels between the objectives of JDL and contextual awareness for general AI is drawn by the authors in \cite{snidaro2019recent}.

Specifically, studies are lacking in the area of data fusion in the NLP space. Most AI models for NLP require a large amount of data to train well, and it is likely that the data comes from different sources, making it similar to a multisensor environment. Moreover, usage of the models beyond the development/testing phase definitely involves data coming in which differs from the original (training) data in terms of source, characteristics etc. Existing data fusion research does not adequately address this area.

In this paper we explore and utilize the synergy between the data fusion tasks and logical interpretable learning of the Tsetlin Machine. We will use TM to assess and interpret the data inconsistencies via logical clauses learned from data. Exploring relationships between the clauses produced by TM we can reveal the data conflicts, that can further be mitigated and resolved. As we will show in experimental study, our framework is quite universal and  can be efficiently applied in various domains, including NLP. To the best of our knowledge, this approach has not been explored in the related work. 

\subsection{Tsetlin Machine Foundation}\label{sec:TM_foundation}
\begin{figure*}[h!]
    \centering
    \begin{subfigure}[t]{\textwidth}
        \centering
        \includegraphics[width=0.6\textwidth]{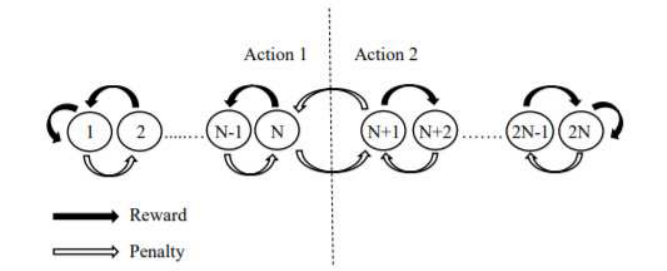}
        \caption{Transition graph of a two-action Tsetlin Automaton}
        \label{figure:2actionta}
    \end{subfigure}%
    \\
    \begin{subfigure}[t]{\textwidth}
        \centering
        \includegraphics[width=0.6\textwidth]{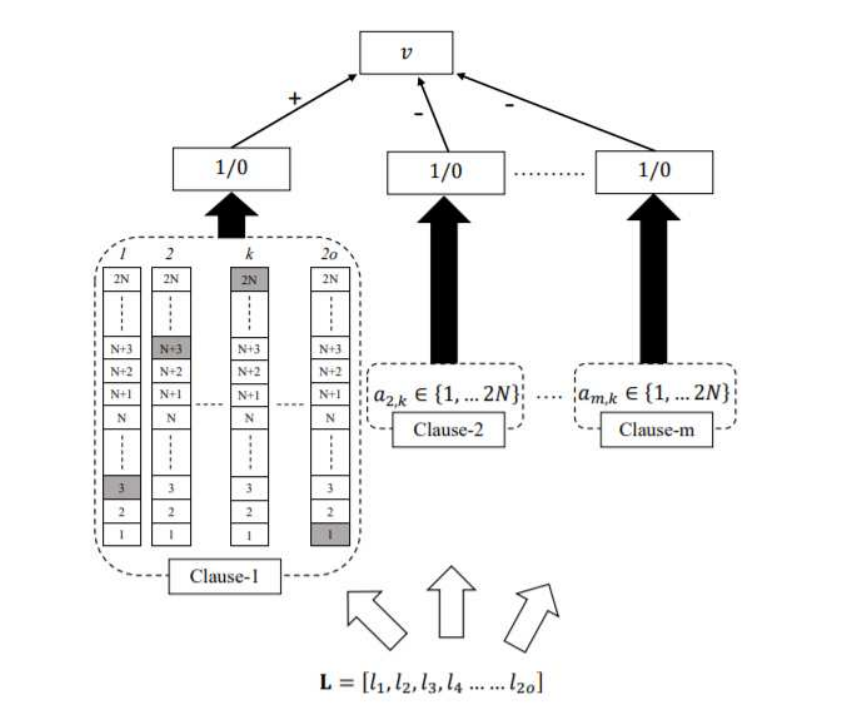}
        \caption{Tsetlin Machine Clause structure}
        \label{figure:tm-clause}
    \end{subfigure}
    \caption{Clause Formation in a Tsetlin Machine}
\end{figure*}

A Tsetlin Automaton (\ac{TA}) is a specialized learning automata that learns to perform the best possible action from the set of actions available for its current state. During training, a reward or penalty is triggered for each action that the \ac{TA} performs, which changes the state. This progressively shifts the \ac{TA} to arrive at the optimal action ~\cite{tsetlin1961behaviour}. Most popularly used is a 2-action \ac{TA}, where a final \ac{TA} state between $1$ and $N$ indicates Action 1, and Action 2 is indicated if the final state is between $N+1$ and $2N$. Figure \ref{figure:2actionta} shows how the reward (and penalty) system works for such a \ac{TA}. As shown, a reward reinforces a particular action by pushing the \ac{TA} state in the direction of that action, whether it is Action 1 or 2, as the case may be. On the other hand, a penalty causes the \ac{TA} state to move towards the central states, i.e. $N$ or $N+1$. 

A group of TAs work in tandem to create a \ac{TM}. In a \ac{TM}, each \ac{TA} is entrusted to keep track of the inclusion or exclusion (what was termed Action 1 and Action 2 in the previous paragraph) of an individual feature. With this mechanism, a \ac{TM} learns a collection of patterns (formed by including or excluding different features in various combinations) that pertain to the task it is performing.

\subsubsection{Classification}

A standard \ac{TM} (also termed as a `regular' \ac{TM} or a Propositional \ac{TM}) requires an input vector $X=(x_1,\ldots,x_f)$ of propositional features, to be classified into one of two classes, $y=0$ or $y=1$. Together with their negated counterparts, $\bar{x}_k\!=\!\lnot x_k = 1-x_k$, the features form a literal set $L\!=\!\{x_1,\ldots,x_f,\bar{x}_1,\ldots,\bar{x}_f\}$.

A \ac{TM} clause is a conjunctive clause $C_j$, formed by ANDing a subset $L_j \subseteq L$ of the literal set:
\begin{equation}
\textstyle
C_j (X)=\bigwedge_{l_k \in L_j} l_k = \prod_{l_k \in L_j} l_k.
\end{equation}
E.g., the clause $C_j(X) = x_1 \land x_2 = x_1 x_2$ consists of the literals $L_j = \{x_1, x_2\}$ and outputs $1$ iff $x_1 = x_2 = 1$.

Each clause has access to $2\times f$ \acp{TA}, one per literal $k$. The bidirectional action of each TA decides which of the literals form a part of a clause. The \ac{TA} states from $1$ to $N$ map to the exclude action, which means that the corresponding literal is excluded from the clause. For states from $N + 1$ to $2N$, the decision becomes include, i.e., the literal is included instead. The states of all the TAs in all of the clauses are jointly stored in the matrix $A: A = (a_{j,k}) \epsilon \{1, \ldots , 2N\}^{m\times2f}$ with $j$ and $k$ referring to the clause and the literal respectively. Thus $I_j^I= \{k|a_{j,k} > N, 1 \leq k \leq 2f\}$ encompasses the literal indices contained in $I_j^I$. Figure \ref{figure:tm-clause} shows the structure of the clauses and the voting scheme diagrammatically.

The number of clauses that the TM should learn per class is a user set parameter~$m$. If there is less information to be learnt than the number of clauses, it leads to empty or repeated clauses. On the other hand, if the number of clauses is not enough, it leads to failure to capture all the information available.  Half of the $m$ clauses are assigned positive polarity ($C_j^+$). The other half is assigned negative polarity ($C_j^-$). The clause outputs, in turn, are combined into a classification decision through summation:
\begin{equation}
\textstyle
v = \sum_{j=1}^{m/2} C_j^+(X) - \sum_{j=1}^{m/2} C_j^-(X).
\end{equation}
The positive clauses vote for accepting the class, i.e. $y=1$ and the negative for rejecting the class, i.e. $y=0$. In other words, Positive clauses capture patterns that are characteristic of the class, where as the other half learns the opposite, i.e. what the patterns should NOT look like if they are to be considered part of the class. Classification is performed based on a majority vote, using the unit step function: $\hat{y} = u(v) = 1 ~\mathbf{if}~ v \ge 0 ~\mathbf{else}~ 0$. The classifier
$\hat{y} = u\left(x_1 \bar{x}_2 + \bar{x}_1 x_2 - x_1 x_2 - \bar{x}_1 \bar{x}_2\right)$, for instance, captures the XOR-relation.

\subsubsection{Learning}

Alg. \ref{algo:tm} details the learning procedure in full. As discussed, learning is performed by a team of $2f$ \acp{TA} per clause, one \ac{TA} per literal $l_k$ (Alg. \ref{algo:tm}, Step~\ref{initialization}). Each \ac{TA} has two actions -- Include or Exclude -- and decides whether to include its designated literal $l_k$ in its clause. 

Step~\ref{inputstep} is where the \ac{TM} processes one training example $(X, y)$ at a time. The \acp{TA} first produce a new random configuration of clauses (Step~\ref{includeexcludestep}), $C_1^+, \ldots, C_{n/2}^-$, followed by Step~\ref{predictstep} where a voting sum $v$ is calculated.

In Steps~\ref{feedbackstart}-\ref{feedbackstop}, based on the difference $\epsilon$ between the clipped voting sum $v^c$, a user-set voting target $T$ and the ground truth, the probability of each \ac{TA} team receiving feedback is calculated. 
The clipped voting sum $v^c$ is used instead of the original $v$, in order to normalize the feedback probability. The voting target for $y=1$ is $T$ and for $y=0$ it is $-T$. Observe that for any input $X$, the probability of reinforcing a clause gradually drops to zero as the voting sum approaches the user-set target. This ensures that clauses distribute themselves across the frequent patterns, rather than missing some and over-concentrating on others. 

The feedback given to the clauses can be divided into two types: Type I feedback which is concerned with producing more frequent patterns, and Type II feedback that is concerned with increasing the discrimination power of the patterns.

\textbf{Type I feedback} is given stochastically to clauses with positive polarity when $y=1$ and to clauses with negative polarity when $y=0$. Each clause, in turn, reinforces its \acp{TA} based on: (1) its output $C_j(X)$; (2) the action of the \ac{TA} -- Include or Exclude; and (3) the value of the literal $l_k$ assigned to the \ac{TA}. Two rules govern Type I feedback:
\begin{itemize}
\item \emph{Include} is rewarded and \emph{Exclude} is penalized with probability $\frac{s-1}{s}$ whenever $C_j(X)=1~\mathbf{and}~l_k=1$. This reinforcement is strong (triggered with high probability) and makes the clause remember and refine the pattern it recognizes in $X$.\footnote{Note that the probability $\frac{s-1}{s}$ is replaced by $1$ when boosting true positives.} 
\item \emph{Include} is penalized and \emph{Exclude} is rewarded with probability $\frac{1}{s}$ whenever $C_j(X)=0~\mathbf{or}~l_k=0$. This reinforcement is weak (triggered with low probability) and coarsens infrequent patterns, making them frequent.
\end{itemize}
Above, the user-configurable parameter $s$ controls pattern frequency, i.e., a higher $s$ produces less frequent patterns.

\textbf{Type II feedback} is given stochastically to clauses with positive polarity when $y=0$ and to clauses with negative polarity when $y\!=\!1$. It penalizes \emph{Exclude} whenever $C_j(X)=1~\mathbf{and}~l_k=0$. Thus, this feedback produces literals for discriminating between $y\!=\!0$ and $y=1$, by making the clause evaluate to $0$ when facing its competing class. For more details on feedback mechanism and the learning process in general, the reader is referred to~\cite{granmo2018tsetlin}.

\begin{algorithm}[t]
\small
\caption{Propositional \ac{TM}}
\label{algo:tm}

\Input{Tsetlin Machine $\mathrm{TM}$, Example pool $S$, Training rounds $e$, Clauses $n$, Features $f$, Voting target $T$, Specificity $s$}

\begin{algorithmic} [1]
\Procedure{Train}{$\mathrm{TM}, S, e, n, f, T, s$}
\For{$j \gets 1, \ldots, n/2$}\label{initialization}
\State $\mathit{TA}_j^+ \gets \mathrm{RandomlyInitializeClauseTATeam}(2f)$
\State $\mathit{TA}_j^- \gets \mathrm{RandomlyInitializeClauseTATeam}(2f)$
\EndFor
\For{$i \gets 1, \ldots, e$}
\State $(X_i, y_i) \gets \mathrm{ObtainTrainingExample}(S)$\label{inputstep}
\State $C_1^+, \ldots, C_{n/2}^- \gets \mathrm{ComposeClauses}(\mathit{TA}_1^+,\ldots,\mathit{TA}_{n/2}^-)$\label{includeexcludestep}
\State $v_i \gets \sum_{j=1}^{n/2} C_j^+(X_i) - \sum_{j=1}^{n/2} C_j^-(X_i)$ \Comment{Vote sum}\label{predictstep}
\State $v_i^c \leftarrow \mathbf{clip}\left(v_i, -T, T\right)$ \Comment{Clipped vote sum}

\For{$j \gets 1, \ldots, n/2$} \Comment{Update \ac{TA} teams}
\If{$y_i = 1$}\label{feedbackstart}
    \State $\epsilon \gets T - v_i^c$ \Comment{Voting error}
    \State TypeIFeedback($X_i, \mathit{TA}_j^+, s$) \textbf{if} rand() $\le \frac{\epsilon}{2T}$
    \State TypeIIFeedback($X_i, \mathit{TA}_j^-$) \textbf{if} rand() $\le \frac{\epsilon}{2T}$
\Else
    \State $\epsilon \gets T + v_i^c$ \Comment{Voting error}
    \State TypeIIFeedback($X_i, \mathit{TA}_j^+$) \textbf{if} rand() $\le \frac{\epsilon}{2T}$
    \State TypeIFeedback($X_i, \mathit{TA}_j^-, s$) \textbf{if} rand() $\le \frac{\epsilon}{2T}$
\EndIf\label{feedbackstop}
\EndFor
\EndFor
\EndProcedure
\end{algorithmic}
\end{algorithm}

 \subsubsection{Weighted Tsetlin Machine}
 Introduced by \cite{phoulady2019weighted}, the vanilla TM can be enhanced by the inclusion of weighted clauses. Since the TM learning is ensemble, it may end up with multiple versions of the same clause (differing in order or positive/negative orientation of literals). The Weighted Tsetlin Machine looks to replace \textit{w} such repeated clauses, with a single copy, with associated \textbf{weight} \textit{w}. Moreover, \textit{w} is also real-valued, allowing for finer control in the feedback process for increasing or decreasing a clause's effect in the overall learning. When the TM is finished training, the weights thus indicate the relative importance of the corresponding clauses to the class under observation.

 \subsubsection{Relational Tsetlin Machine}
The Tsetlin Machine was further supplemented to take advantage of preexisting logical structures present in data via the introduction of the Relational Tsetlin Machine (RTM) \cite{saha2022relational}. The RTM processes relations, variables and constants with the aim of building a set of Horn Clauses from the clauses learnt during training. These Horn Clauses are shown to describe a logical estimation of the world the TM is operating in.

\section{Proposed Approach}
\label{sec:Proposal}

\subsection{Using Tsetlin Machine for Assessing and Fusing Dynamic Data }

One of the major advantages of the Tsetlin Machine, as highlighted by \cite{yadav2021human, berge2019using, saha2022relational, zhang2023interpretable} is its explainability. Most existing research using TMs have focused on exploring how the decision can be bolstered by using the TM's clauses as explanations for the decisions. 

As described in the previous section, a TM learns clauses based on repetitive and discriminative patterns occurring in the training data. We theorize that since the decision has been demonstratively explained by clauses, the decision making process itself can also be leveraged in ways to understand more about the data. Even the interaction between the data and the model can be explored by following the TM's decision making.

In \cite{saha2020mining}, the authors explain how the learnt clauses are the global description of the class. An individual example, say from the test set, will not match all of the patterns encoded in the clauses of its (the example's) class. Only certain clauses from each possible class are triggered in the presence of an individual sample. The decision making process is democratic, the triggered clauses vote either for (if they are positive clauses) or against (negative clauses) a particular class, and the class with most votes wins in the end. 

Given the above, we hypothesize that differences in data can be picked up by observing differences in TM learning over the different data.

Given an initial data, we assume that we can consider the data as clean, since we do not have any other information on the data that may suggest otherwise. Therefore, any further additions to this data, which is not consistent with the characteristics of the initial data, is considered noise. One can argue that any particular `inconsistent characteristic' can be regarded as noise only until there is sufficient evidence of its occurrence amongst the entirety of the data. Given sufficient evidence, an `inconsistency' becomes a characteristic of the data itself.

To begin, we show that a TM's learning is consistent.  That is, when tested with a particular test sample, different TMs trained on the same data,  gives (a) the same classification decision and (b) the same local explanation for the data. This is to ensure that when we compare two versions of TM learning, we observe the same differences (or similarities) between the two versions every time. Two different versions of TM learning can be from differences in the input data or from differences in the learning parameters. In the experiments described subsequently, we use differing input data to achieve different versions of learning by TMs with same learning parameters.

\begin{figure*}[h!]
\centering
\includegraphics[width=0.8\textwidth]{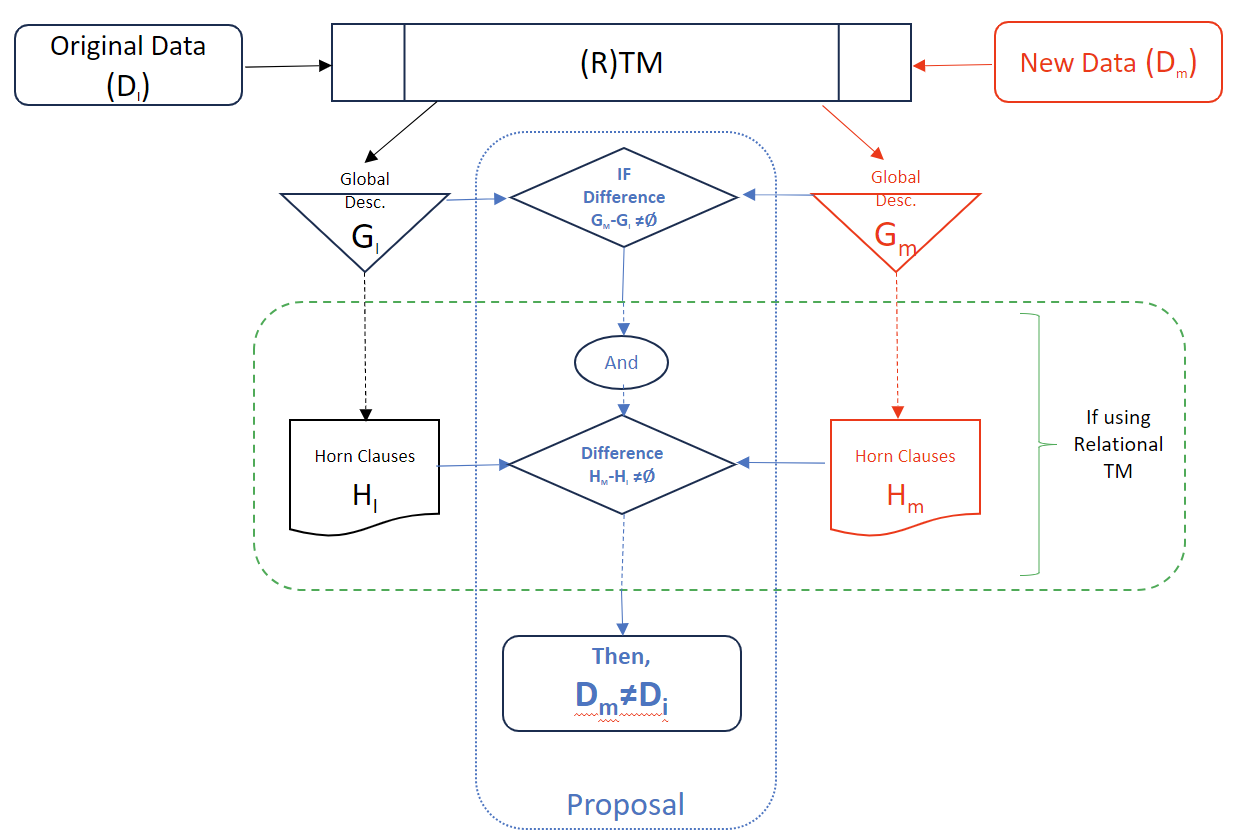}
\caption{Proposed approach for tracking changes in data}
\label{figure:proposal}
\end{figure*}

The rest of the proposed approach is described herewith, together with Figure \ref{figure:proposal}.
Thus given the initial data ($D_I$), the TM trained on it accurately captures the global characteristics of $D_I$. Let this description of I be termed as $G_I$. Using a RTM, we can base a set of Horn clauses $H_I$ on $G_I$ to give us a logical view of the world. This is shown in the left column of Figure \ref{figure:proposal}.

Next, as shown via the right most column of Figure \ref{figure:proposal}, we have some new data, $D_m$, which differs from $D_I$ by an amount $m\%$. In other words, $m$ number of samples out of every 100 differ in some aspect from the original $D_I$. This subset of the data, denoted by $D_{m} - D_{I}$, is thus inconsistent with the global description $G_I$ of $D_I$. The global description arising from $D_m$ is termed $G_m$. When we do not have any prior knowledge of the nature of the differentiation between $D_I$ and $D_m$, we expect to obtain a good approximation from $G_m - G_I$. this is the cornerstone of our approach, and is illustrated in the middle column of Figure \ref{figure:proposal}.



Moreover, if the set of Horn Clauses obtained from the initial data via the RTM ($H_I$) shows us a recursive logical definition of the world, the equivalent Horn Clauses from the modified data ($H_m$) can define an almost-recursive world. Conversely, removing certain samples from $D_m$ would enable us to arrive at completely recursive set of Horn Clauses. 

We perform experiments on a generated dataset (explained in section \ref{subsec:Synthetic Experiments}) to highlight how changes in data are visible in learning.

\subsection{Illustrating the Proposed Method with a Synthetic Example }
\label{subsec:Synthetic Experiments}

We will illustrate the proposed approach using synthetic data, specifically designed to highlight aspects of our proposal. 
Inspired by the Scene domain in the SCONE dataset \cite{long2016simpler}, this involves a simple set of texts. The universe consists of $n$ persons located next to each other, and one hat. The accompanying text consists of $m$ sentences, each sentence describing the current owner of the hat deciding to either (i) pass the hat left (if possible) or (ii) pass the hat right (if possible) or (iii) do nothing. The task is to identify who has the hat last. Currently we limit the data to a maximum of 4 distinct persons and 3 consecutive actions.

We generate 10000 samples of this data, and hereafter refer to it as 'Hat-Passing Data'.

{\bf Sample Data:}
As an example, we have a scenario as shown in Figure \ref{fig:data three images}.  The original cap owner, as in Figure \ref{fig:data three images}$>$Start State, is implicit. 
The set of actions is described to the model as per the following text: 
\textit{B passes the hat right. C passes the hat left. B passes the hat left.} [See Fig.\ref{fig:data three images}$>$ Textual Information]
The task answer (i.e. the answer to the question, 'Who has the hat last?'), as shown in Figure \ref{fig:data three images}$>$Final State, is Person A. 

\setlength{\tabcolsep}{1pt}
\begin{table*}[h!]
 \centering     
\begin{tabular}{|c|c|c|}
\hline
\begin{minipage}{.3\textwidth}\centering      \includegraphics[width=0.8\linewidth, scale=0.6]{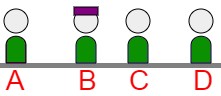}  \label{fig:hats_start}  \end{minipage}                  & \multirow{2}{*}{\begin{minipage}{.3\textwidth} \centering     \includegraphics[width=0.7\linewidth]{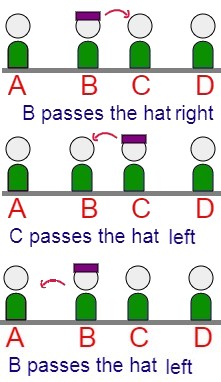} \label{fig:hats_text_info}   \end{minipage}} &             \\
\multicolumn{1}{|l|}{} &                        & \begin{minipage}{.3\textwidth} \vspace{8em} \centering   \includegraphics[width=0.8\linewidth, scale=0.5]{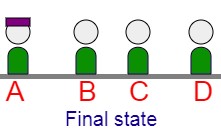} \label{fig:hats_answer}   \end{minipage}         \\ \hline
Start State            & Textual Information    & Final State \\ \hline
\end{tabular}
\caption{One example from experimental data, showing the (implicit) start state, the movements and finally the end state (answer).}
\label{fig:data three images}
\end{table*}

\par

{\bf Relational setup:}
Each sentence in the text is represented as a tuple (Person, Action). Person can take values [A,B,C,D], representing names. Action can take values [R,L,N] representing Right, Left and Nothing respectively. To indicate the sentence order, we also include the time instant, making the final tuples of the form (Tx\_Person,Action). (Tx) represents the $x^{th}$ time instant, and Person,Action represents the CapOwner at the $x^{th}$ time instant, and the action chosen by them.
The example described above (Fig. \ref{fig:data three images}) is thus represented as the following in the relational setup:

(T0\_B,L), (T1\_C,R), (T2\_B,R)

From the perspective of the Tsetlin Machine based classification system, the system is given the above representation as input. It predicts the answer to the question 'Who has the cap last?' by learning to classify between the classes [A,B,C,D] (i.e. the set of persons) based on the input.

{\bf Consistency in learning:}
The first experiment we perform on this dataset is to establish that the Tsetlin Machine learns the same patterns consistently. Since our proposal hypothesizes that change in data can be traced by observing the change in learning, first we confirm that when the data is constant, so is the associated learning.

Using the `Hat-Passing' data, we obtain the following results. At the global-level , 98\% of clauses were built from the same features over 100 individual runs. At the local level, in 97.2\% of the test samples, the TM used the same clausal reasoning to make the decision. This confirms that the TM learns similarly across runs.

\subsubsection{Horn Clauses on Initial Data}

As mentioned previously, the system is tasked with classifying the relational representation of the text into one of the possible classes, as an answer to the task question. Table \ref{table-initial-clause-hat} shows a consolidated view of the clauses learnt for the 4 classes in the experimental setup.
\label{section-shortened-horn-clauses}

\setlength{\tabcolsep}{1pt}
\begin{table}[h!]
\centering
\begin{tabular}{|lll|}
\hline
\multicolumn{1}{|c|}{\textbf{Type\#}} &
  \multicolumn{1}{c|}{\textbf{Clause}} &
  \multicolumn{1}{c|}{\textbf{Wt}} \\ \hline
\multicolumn{3}{|c|}{Class: Person A} \\ \hline
\multicolumn{1}{|l|}{$+ve Cl.\#0$} &
  \multicolumn{1}{l|}{$T2\_A,N$} &
  25 \\ \hline
\multicolumn{1}{|l|}{$+ve Cl.\#1$} &
  \multicolumn{1}{l|}{$T2\_B,L$} &
  16 \\ \hline
\multicolumn{1}{|l|}{$-ve Cl.\#0$} &
  \multicolumn{1}{l|}{$T2\_B,N \land \neg T2\_B,L$} &
  -13 \\ \hline
\multicolumn{1}{|l|}{$-ve Cl.\#1$} &
  \multicolumn{1}{l|}{$\neg T2\_A,N \land \neg T2\_B,L$} &
  -20 \\ \hline
\multicolumn{3}{|c|}{Class: Person B} \\ \hline
\multicolumn{1}{|l|}{$+ve Cl.\#0$} &
  \multicolumn{1}{l|}{$T2\_A,R$}  &
  15 \\ \hline
\multicolumn{1}{|l|}{$+ve Cl.\#1$} &
  \multicolumn{1}{l|}{$T2\_B,N$} &
  15 \\ \hline
\multicolumn{1}{|l|}{$-ve Cl.\#0$} &
  \multicolumn{1}{l|}{$\neg T2\_C,R \land \neg T2\_B,N$} &
  -35 \\ \hline
\multicolumn{1}{|l|}{$-ve Cl.\#1$} &
  \multicolumn{1}{l|}{$T2\_C,N$} &
  -5 \\ \hline
\multicolumn{3}{|c|}{Class : Person C} \\ \hline
\multicolumn{1}{|l|}{$+ve Cl.\#0$} &
  \multicolumn{1}{l|}{$\neg T2\_C,L \land T2\_C,N$} &
  15 \\ \hline
\multicolumn{1}{|l|}{$+ve Cl.\#1$} &
  \multicolumn{1}{l|}{$T2\_D,L$} &
  12 \\ \hline
\multicolumn{1}{|l|}{$-ve Cl.\#0$} &
  \multicolumn{1}{l|}{$T2\_C,R$} &
  -27 \\ \hline
\multicolumn{1}{|l|}{$-ve Cl.\#1$} &
  \multicolumn{1}{l|}{$\neg T2\_B,R \land \neg T2\_C,N \land \neg T2\_D,L $} &
  -33 \\ \hline
\multicolumn{3}{|c|}{Class : Person D} \\ \hline
\multicolumn{1}{|l|}{$+ve Cl.\#0$} &
  \multicolumn{1}{l|}{$T2\_C,R \land \neg T2\_A,N$} &
  25 \\ \hline
\multicolumn{1}{|l|}{$+ve Cl.\#1$} &
  \multicolumn{1}{l|}{$T2\_D,N$} & 
  20 \\ \hline
\multicolumn{1}{|l|}{$-ve Cl.\#0$} &
  \multicolumn{1}{l|}{$\neg T2\_D,N$} &
  -9 \\ \hline
\multicolumn{1}{|l|}{$-ve Cl.\#1$} &
  \multicolumn{1}{l|}{$\neg T2\_C,R \land \neg T2\_D,N$} &
  -15 \\ \hline
\end{tabular}
\caption{Representative clauses for each class for the `Hat-Passing' data.}
\label{table-initial-clause-hat}
\end{table}

Based on the clauses that the TM learns, we can arrive at the following Horn Clauses :
\begin{itemize}
    \item[1.] CapOwner(X)

\item[2.] CapOwner(Y) $<-$ CapOwner(X), PassCap(X,Y)

\item[3.] PassCap(X,Y) $<-$ RightNeighbour(X,Y) or LeftNeighbour(X,Y)

 \item[4.] RightNeighbour(X,Y) $<-$ LeftNeighbour(Y, X)
\end{itemize}

These Horn clauses capture the behavioural semantics of the world as described by the text over multiple training samples.
 \noindent
 Using the terminology introduced in Section \ref{sec:Proposal}, this set of Horn Clauses is thus $H_I$, based on $D_I$. Comparing the global description and the Horn Clauses, we see that the Horn Clauses capture a recursive relationship. While the global description, i.e. all the clauses learnt by the TM, form a verbose description of the world the classifier is tasked with understanding, the Horn Clauses capture a more refined (or succint) representation of the same. 

 \subsubsection{Targeted Destabilization}
 \label{targetted-destabilization}

To explore how the Tsetlin Machine reacts to perturbations in the data, we introduce some targeted destabilization based on the Horn Clauses.

In other words, we introduce deviations that are specifically non-conforming to certain Horn Clauses obtained from $D_I$. The presence of such deviations will purportedly `destabilize' the learning. 

To illustrate the idea better, we continue with the same example as above. As seen in the section \ref{section-shortened-horn-clauses}, the Horn Clauses $H_I$ obtained from $D_I$ reaches the conclusion  PassCap(Y,X) $<-$ Neighbour (X,Y) and CapOwner (Y). In natural language, it summarizes that a cap can be passed to a neighbour of the current cap owner. 

In our experiments, we introduce examples that contradict this conclusion. Therefore, we add some examples, where Person A being the cap owner, passes right, and the consequent cap owner is Person C, who is not a neighbour (instead of Person B, who is a neighbour of Person A). When the TM learns from this new dataset $D_m$, it includes certain clauses to capture the contradiction, and hence the resultant Horn Clauses $H_m$, do not conform to $H_I$ any more. 

As a next step, even more inconsistency is added. This time, it is of the form of removal of some examples, where cap owner B passed the cap right, resulting in Person C becoming the cap owner. This further jeopardizes the conclusion drawn in $H_I$ that a cap can be passed to a cap owner`s neighbour. 
\begin{itemize}
    \item Destabilizing \textit{PassCap(Y,X) $<-$ Neighbour(X,Y)}

  We design a system to verify the above Horn Clause. Given a `PassCap' information between two persons, we try to determine if two persons are neighbours. The pass information is encoded as Pass[A,B] to represent person A passing the cap to person B. The determination part is encoded as binary by framing it as a query (is person A and person B neighbours, represented as Query\textunderscore~IsNeighbour[A,B] in the TM). The answer to the query is thus binary (Yes/No).

 Given the initial experimental data, the resultant clauses of this task for Class 'Yes' are as follows (the number following the clause is the weight of the clause):

Pass[A,B] or Pass[B,A] and Query\textunderscore IsNeighbour[A,B] : 43

Pass[B,C] or Pass[C,B] and Query\textunderscore IsNeighbour[B,C] : 12

Pass[C,D] or Pass[D,C] and Query\textunderscore IsNeighbour[C,D] : 14

Next we set some examples where Pass[A,B] is false, but the corresponding Query\textunderscore IsNeighbour[A,B] is still set to Yes. That is, a constructed scenario where A cannot pass to B, but A and B are neighbours.

 At this point,  the resultant clauses (again for Class 'Yes') look like :

Pass[A,B] or Pass[B,A] and Query\textunderscore IsNeighbour[A,B] : 16

Pass[B,C] or Pass[C,B] and Query\textunderscore IsNeighbour[B,C] : 13

Pass[C,D] or Pass[D,C] and Query\textunderscore IsNeighbour[C,D] : 12

If we further increase the incidence of this scenario that does not conform to the original world-view, we observe that weight of the affected clause drops further. Consequently, there is less evidence in support of the original Horn Clauses (subsec \ref{section-shortened-horn-clauses}).

\item Destabilizing \textit{Neighbour(X,Y)  $<-$  PassCap(Y,X)}

Similarly, we set up an experiment to test the reflexive Horn Clause. 
Given information that person A and B are neighbours, we determine if a PassCap is allowed between them.  The task is setup such that the neighbour information is of the form Neighbour[A,B]. The pass to be determined as valid or invalid is included as a query, Query\textunderscore IsValidPass[A,B]. The output is thus Yes/No, depending on whether the pass is valid given the neighbour information. According to the original data, this 2 persons are neighbours, then a pass between them is always valid. This leads to the following clauses for Class 'Yes' (the number stated at the end is the weight of the clause):

Neighbour[A,B] and Query\textunderscore IsValidPass[A,B] or \\Query\textunderscore IsValidPass[B,A] : 15

Neighbour[B,C] and Query\textunderscore IsValidPass[B,C] or \\Query\textunderscore IsValidPass[C,B] : 15

Neighbour[C,D] and Query\textunderscore IsValidPass[D,C] or \\Query\textunderscore IsValidPass[C,D]: 13

Here we then set some examples where Neighbour[A,B] is false, but the corresponding Query\textunderscore IsValidPass[A,B] or Query\textunderscore IsValidPass[B,A] still have output Yes. 

In this scenario, we do not observe a drop in weight as in the previous case. Instead, we find that the `affected' clause gets mangled, and does not retain the clear Horn-clause-compliant form seen in the original data. The unaffected clauses remain same, indicating that it is the introduced changes in the data that resulted in the changes in the clauses. However, this `mangled' clauses are not standard across different TM trainings with the same data. Given below is one example of how the clauses differ from the global view obtained from the original data by one clause.

Class Yes:

Not~Neighbour~[A,B] and Neighbour~[B,C] and Query\textunderscore~IsValidPass~[A,B] or \\Query\textunderscore IsValidPass~[B,A] : 10

Neighbour[B,C] and Query\textunderscore IsValidPass[B,C] or \\Query\textunderscore~IsValidPass[C,B] : 16

Neighbour[C,D] and Query\textunderscore IsValidPass[D,C] or \\Query\textunderscore IsValidPass[C,D]: 18
\end{itemize}

Similar to the previous case, the current scenario also results in lower evidence in support of the original Horn Clauses (subsec \ref{section-shortened-horn-clauses}). With the new malformed clauses, it is no longer possible to arrive at the recursive Horn clauses that we saw from the original data. Instead, the resultant Horn Clauses here are recursive only for certain cases (cases other than for Neighbours A,B, which were deliberately targetted here), leading to {\em an almost-recursive} set of Horn Clauses.

 \subsubsection{Non-targeted Destabilization}
 \label{nontargetted-destabilization}

 Next we explore how the Tsetlin Machine reacts in a more organic scenario. In the previous section, we used the information about $H_I$ to construct our perturbations. Here we try to use our knowledge of the world (without involving the Horn Clauses) to introduce disturbances in the data, and see how the learning of the TM is affected.

 As an example, we introduce 5-30\% inconsistencies in the `Hat-Passing' data. In the experiment, first a random selection of input samples is collected, all of which either have A and/or D (the first and the last person in the row) as the cap owner at some point. According to the initial worldview, A does not have a LeftNeighbour and D does not have a RightNeighbour. Consequently, PassLeft is an invalid action for A, and similarly, PassRight is invalid for D. The selected input samples all have valid actions, according to the initial worldview as described by the Horn clauses. Now, to introduce inconsistensies, a valid movent by A (either PassRight or DoNothing) is replaced by the invalid action PassLeft. Similar modification is done to movements by D in the selected inputs. 
\begin{table}[h!]
\centering
\begin{tabular}{|lll|}
\hline
\multicolumn{1}{|c|}{\textbf{Type\#}} &
  \multicolumn{1}{c|}{\textbf{Clause}} &
  \multicolumn{1}{c|}{\textbf{Wt}} \\ \hline
\multicolumn{3}{|c|}{Class: Person A} \\ \hline
\multicolumn{1}{|l|}{$+ve Cl.\#0$} &
  \multicolumn{1}{l|}{$T2\_A,N$} &
  15 \\ \hline
\multicolumn{1}{|l|}{$+ve Cl.\#1$} &
  \multicolumn{1}{l|}{$T2\_B,L$} &
  15 \\ \hline
\multicolumn{1}{|l|}{$-ve Cl.\#0$} &
  \multicolumn{1}{l|}{$\neg T2\_A,N \land \neg T2\_B,L \land \mathbf{\neg T2\_A,L}$} &
  -29 \\ \hline
\multicolumn{1}{|l|}{$-ve Cl.\#1$} &
  \multicolumn{1}{l|}{$\neg T2\_A,N \land \neg T2\_B,L$} &
  -21 \\ \hline
\multicolumn{3}{|c|}{Class: Person B} \\ \hline
\multicolumn{1}{|l|}{$+ve Cl.\#0$} &
  \multicolumn{1}{l|}{$T2\_A,R$}  &
  13 \\ \hline
\multicolumn{1}{|l|}{$+ve Cl.\#1$} &
  \multicolumn{1}{l|}{$T2\_B,N$} &
  13 \\ \hline
\multicolumn{1}{|l|}{$-ve Cl.\#0$} &
  \multicolumn{1}{l|}{$\neg T2\_C,R \land \neg T2\_B,N$} &
  -30 \\ \hline
\multicolumn{1}{|l|}{$-ve Cl.\#1$} &
  \multicolumn{1}{l|}{$T2\_C,N$} &
  -24 \\ \hline
\multicolumn{3}{|c|}{Class : Person C} \\ \hline
\multicolumn{1}{|l|}{$+ve Cl.\#0$} &
  \multicolumn{1}{l|}{$\neg T2\_C,L \land T2\_C,N$} &
  13 \\ \hline
\multicolumn{1}{|l|}{$+ve Cl.\#1$} &
  \multicolumn{1}{l|}{$T2\_D,L$} &
  16 \\ \hline
\multicolumn{1}{|l|}{$-ve Cl.\#0$} &
  \multicolumn{1}{l|}{$T2\_C,R$} &
  -17 \\ \hline
\multicolumn{1}{|l|}{$-ve Cl.\#1$} &
  \multicolumn{1}{l|}{$\neg T2\_B,R \land \neg T2\_C,N \land \neg T2\_D,L$} &
  -13 \\ \hline
\multicolumn{3}{|c|}{Class : Person D} \\ \hline
\multicolumn{1}{|l|}{$+ve Cl.\#0$} &
  \multicolumn{1}{l|}{$T2\_C,R \land \neg T2\_A,N$} &
  15 \\ \hline
\multicolumn{1}{|l|}{$+ve Cl.\#1$} &
  \multicolumn{1}{l|}{$T2\_D,N$} & 
  14 \\ \hline
\multicolumn{1}{|l|}{$-ve Cl.\#0$} &
  \multicolumn{1}{l|}{$\mathbf{\neg T2\_D,R}$} &
  -21 \\ \hline
\multicolumn{1}{|l|}{$-ve Cl.\#1$} &
  \multicolumn{1}{l|}{$\neg T2\_C,R \land \neg T2\_D,N$} &
  -30 \\ \hline
\end{tabular}%

\caption{Representative clauses for each class for the `Hat-Passing' data with 15\% included inconsistencies. Deviations from Table\ref{table-initial-clause-hat} is in bold.}
\label{table-secondary-clause-hat}
\end{table}

Table \ref{table-secondary-clause-hat} shows how the global view by the clauses changes accordingly. The deviations from the previously learnt clauses (\ref{table-initial-clause-hat}) are marked in bold. It is seen that the two sets of learnt clauses differ exactly by the inconsistencies introduced in this step.

\subsubsection{Acknowledgement of change}
Subsection \ref{targetted-destabilization} and Subsection \ref{nontargetted-destabilization} are the basis for our claim that TM learning can be used to acknowledge the presence of changes in data with respect to a baseline. Comparing the two global views of clauses constructed on two datasets, one can determine if there are differences in the characteristics of the data in them.

\subsubsection{Identification of inconsistencies}
\label{subsec:synthetic inconsistencies}
Finally, we propose a method of identifying and isolating the deviations to some extent from $D_{m}$. We have previously mentioned how the global view of the learning is characteristic of the data on which the learning is performed. We divide $D_m$ into multiple overlapping segments or cuts ($D_{ci},$), each randomly generated and of a fixed size, smaller than the total size of $D_m$. $D_c = {D_{c1}, D_{c2} ... D_{cn}.}$ We train TM with same parameters as above, on each of $D_{ci}$ and obtain a global description $G_{ci}$. We compare $G_{ci}$ with $G_I$, and assign the comparison a numerical value ($V_{ci}$) calculated on the basis of overlap amongst the individual clauses. Once the entire set $D_{c}$ is thus processed, the values in $V_{c}$ ($= {V_{c1},V_{c2}...V_{cn}}$) are sorted in a ascending manner. 
The idea behind this step is that the cut(s) with maximum overlap are the ones which are closest to $D_I$ characteristically. Subsequently, $m$ segments which have the least $V_{ci}$ are chosen and form a trial removal set $R$. $R_j$ is removed from $D_m$ one at a time, and the resultant learning is again compared with $G_I$. If the new comparative value $VR_{j}$ is greater than the comparative value of $G_I$ and $G_m$, we can conclude that $R_j$ contains at least some data which does not follow the general description of $D_I$. In other words, if $D_m-R_j$ (or $D_{R_j}$ in short) is closer to $D_I$ than $D_m$, then $R_j$ contains some characteristics that are not seen in $D_I$, and may be worth investigating. The number of cuts $n$, or the number of items in the trial removal set $m$ both will depend on the larger task at hand. 

It is important to note that the numerical value assigned to the overlap between learnings does not have much significance just by itself. It is merely used to judge comparative overlap between learning obtained from different data sources.

{\bf Mitigating or Amplifying Inconsistencies:}
What is to be done with the cuts $R_j*$ that are identified to be significantly deviant, is also left to be task dependant. In certain scenarios, it maybe relevant to get rid of that data altogether. It is also possible to train an additional Tsetlin Machine to handle such deviant data. Leaving the data as is of course possible, and knowledge of the difference in data characteristics may be used else where.

As an example, in one experiment, the comparative value between $D_I$ and $D_m$ was 0.78 ($V_{I-m}$). With $n$=10 cuts, the obtained $V_{c}$ were in the range 0.7445 to 0.8797. Removing the cuts that had the $5$ least values of $V_{c}$ lead to the comparative value of $V_{I-R_j}$ greater than $V_{I-m}$ by 0.03 to 0.15. Of these, the cuts corresponding to the greatest amount of increase were verified to have more number of data that is inconsistent with the original data.

\section{Experimental Study}
\label{sec:Real Experiments}
In this section, we conduct experiments on three different datasets to highlight how our proposed method performs in real world scenarios, and what modifications and/or supplements are necessary.

\subsection{Experiments with Network Intrusion Data}
\label{subsec:Network Experiments}

The CICIDS2017 dataset \cite{sharafaldin2018toward}  consists of labeled network flows designed and collected with the aim of creating a dataset with a wide variety of possible network intrusion attacks that can be identified by their characteristics via machine learning. 

\begin{table*}[h]
\centering
\begin{tabular}{|l|l|l|}
\hline
\textbf{Notation used here} & \textbf{Feature Name} & \textbf{Feature Description} \\ \hline
Init\_Win\_bytes\_bwd\_(x::y] & Init\_Win\_bytes\_backward & \begin{tabular}[c]{@{}l@{}}The total number of bytes sent in initial window in the\\ backward direction\end{tabular} \\ \hline
Bwd\_IAT\_Min\_(x::y] & Bwd IAT Min & \begin{tabular}[c]{@{}l@{}}Minimum time between two packets sent in the \\ backward direction\end{tabular} \\ \hline
Fwd\_IAT\_Std\_(x::y] & Fwd IAT Std & \begin{tabular}[c]{@{}l@{}}Standard deviation time between two packets sent in the \\ forward direction\end{tabular} \\ \hline
Init\_Win\_bytes\_fwd\_(x::y] & Init\_Win\_bytes\_forward & \begin{tabular}[c]{@{}l@{}}The total number of bytes sent in initial window in the \\ forward direction\end{tabular} \\ \hline
Active\_Mean\_(x::y] & Active Mean & Mean time a flow was active before becoming idle \\ \hline
Idle Min\_(x::y] & Idle Min & Minimum time a flow was idle before becoming active \\ \hline
Flow\_Duration\_(x::y] & Flow duration & Duration of the flow in Microsecond \\ \hline
Idle\_Mean\_(x::y] & Idle Mean & Mean time a flow was idle before becoming active \\ \hline
Subflow\_Bwd\_Bytes\_(x::y] & Subflow Bwd Bytes & \begin{tabular}[c]{@{}l@{}}The average number of bytes in a sub flow in the\\ backward direction\end{tabular} \\ \hline
Bwd\_Packets/s\_(x::y] & Bwd Packets/s & Number of backward packets per second \\ \hline
Flow\_IAT\_Std\_(x::y] & Flow IAT Std & \begin{tabular}[c]{@{}l@{}}Standard deviation time between two packets sent\\ in the flow\end{tabular} \\ \hline
\end{tabular}
\caption{Details of features in Network Intrusion Data}
\label{tab:NetworkIntrusionDatafeatures}
\end{table*}

The feature-set comprises of more  80 network traffic features. In order to binarize the data, the features were binned, with every bin consisting of 10 percentile of the range of values taken by that feature. Table \ref{tab:NetworkIntrusionDatafeatures} shows some of these features. The first column mentions the notation used in this work, and each feature is followed by ``\_(x::y]'', which indicates the limits of the bin. the second and third columns mention the feature name and the feature description as in the CICFlowMeter documentation \cite{engelen2021troubleshooting}.

The data is collected over the period of a work week. Monday`s data consisted of only benign traffic. Tuesday's consisted of benign traffic as well as <<Specific attack>> traffic, which were of two types, viz. <<TuesdayAttack1>> and <<TuesdayAttack2>>. Similarly, Wednesday, Thursday and Friday all had both benign and attack traffic. On Wednesday, the attack traffic was of 3 types : <<WednesdayAttack1>>, <<WednesdayAttack2>> and <<WednesdayAttack3>>. 


In the first set of experiments, we consider a binary classification problem.
A TM classification model, termed as `TMTuesday', is initially trained to differentiate between 2 classes : Benign and Malicious\_Tuesday.
Similarly, a separate TM, termed as `TMWednesday', is trained to differentiate between 2 classes : Benign and Malicious\_Wednesday.

Comparing the clauses learnt for the two days we observe the following:


\begin{itemize}
    \item Clauses learnt for `Malicious\_*' class differ significantly between `TMTuesday' and `TMWednesday' (Similarity as calculated previously is at 4.38)

    \textbf{Examples for class Malicious for TMTuesday:} 
\begin{itemize}
    \item Init\_Win\_bytes\_bwd\_(374341.95::392167.72] AND \\Bwd\_IAT\_Min\_(124781.32::142607.08]
    \item Fwd\_IAT\_Std\_(1.0::17826.76]
\end{itemize}
    
    \textbf{Examples for class Malicious for TMWednesday: }

    \begin{itemize}
    \item Init\_Win\_bytes\_fwd\_(124781.32::142607.08] \\ AND NOT Active\_Mean\_(1.0::17826.76]
    \item Idle Min\_(427819.24::445645.0] AND \\Idle\_ Min\_(124781.32::142607.08]
\end{itemize}

    \item Clauses learnt for `Benign' class are somewhat similar (58.9)

    \textbf{Examples for class Benign for TMTuesday: }

    \begin{itemize}
        \item NOT Fwd\_IAT\_Std\_(1.0::17826.76]
        \item Flow\_Duration\_(409993.48::427819.24] AND \\NOT Idle\_Mean\_(142607.08::160432.84]
    \end{itemize}

     \textbf{Examples for class Benign for TMWednesday:} 

     \begin{itemize}
        \item NOT Fwd\_IAT\_Std\_(1.0::17826.76] AND NOT Subflow\_Bwd\_ Bytes\_(1.0::17826.76]
        \item Bwd\_Packets/s\_(1.0::17826.76] AND \\Idle\_~Mean\_(1.0::17826.76] AND \\NOT~Flow\_IAT\_Std\_(1.0::17826.76]
    \end{itemize}

\end{itemize}

This can be attributed to the fact that in each case, the TM built a definition of Benign with the perspective of the difference in Malicious traffic on Tuesday and Wednesday.

The second set of experiments treat the sub-types of attacks as separate classes, as compared to the Benign vs Malicious approach stated above.
A TM classification model, termed as `TMTuesday', is trained to differentiate between 3 classes : Benign, <<TuesdayAttack1>> and <<TuesdayAttack2>>.
Similarly, a separate TM, termed as `TMWednesday', is trained to differentiate between 4 classes : Bening, <<WednesdayAttack1>>, <<WednesdayAttack2>> and <<WednesdayAttack3>>.

For obvious reasons, the only direct comparison can be between the clauses learnt for class Benign by `TMTuesday' and those learnt for class Benign by `TMWednesday'. As before, although both sets of clauses define benign traffic, we observed considerable difference between the two. Since the perspectives of <<TuesdayAttack>> and <<WednesdayAttack>> are different, so are the global views of Benign traffic according to `TMTuesday' and those learnt for class Benign by `TMWednesday'. 

In both experiments, the difference in learnt clauses can be used to indicate the presence of a new information, which, in this case, is a new form of malicious traffic.

\subsubsection{Data fusion task}
To explore this data for a data fusion task, we append the data from Tuesday to the data from Wednesday, and so on. This can be considered as a binary classification task.

Next we create `TMTuesday' and `TMTues+Wed' and compare the learnt clauses.
Firstly, the clauses overall are different (very low similarity), which should allow us to acknowledge that the data has changed. In this case, though the TM is tasked with simply understanding <<Benign + Malicious>>, the data changes from <<Tuesday Attack + Benign>> to <<TuesdayAttack + WednesdayAttack + Benign>>.
For the Benign Class, while the clauses learnt were different, the ones that were similar had very high weight.
For example, ``NOT Active\_Min\_(1.0::17826.76]'' was a clause common to the different cases, and had a weight of >150 in both TMs. Other non-common clauses had weights <100 in both TMs.

When working with synthetic data, we had the advantage of being able to control the data complexity. As a result, the experiments conducted on the Hat-Passing data showed learnt clauses that captured all the semantics that the data had to offer. When new (modified) data was added, the newly learnt clauses could be compared with the older clauses. Real data, on the other hand, is more complex. The clauses that the Tsetlin Machine learns often focuses only on some of the information that the data provides and the semantic aspect captured depends on the comparative examples that the machine is provided with. As new data is added in this scenario, the clauses can no longer be directly compared since they may now choose to focus on some other aspect (even of the old data) in order to successfully carry out the classification task on the combined (new+ old) data.

Therefore when the data complexity is high, we need to refer to parameters such as clause weight (as discussed here) or average difference of clause sums ( discussed subsequently) in order to determine change in data.

\subsubsection{Using Average Sum Difference as a Metric}
Another metric for identifying if data is compatible with existing learning is obtaining the average sum difference (ASD) from the TM. Once training is completed, the TM classifies a sample by having clauses vote either for or against each class. The predicted class is the class which get most votes, thresholded by the parameter T (details in section \ref{sec:background}). ASD is calculated as the absolute difference between the weighted clause sum for Class 1 and the weighted class sum for Class 0 (without the thresholding used in the prediction mechanism). Similar to the overlap score used in Subsection \ref{subsec:synthetic inconsistencies}, ASD is also not an absolute measure, and cannot be used to compare between different TM setups. On the same setup however, testing on similar data tends to result in similar ASD values. A higher ASD indicates a more 'confident' decision on average, compared to a scenario where the TM gives a lower ASD on the testing data. Hence ASD can be used to indicate whether the existing learning is good enough for the incoming data.

For example, training 'TMTuesday1' on half of Tuesday attack data, and 'TMTuesday2' on the entirety of Tuesday attack data, gave ASD values as 1236.159 and 1265.5753 respectively. In both cases the testing data was 20\% of the available data, and the rest was for training. Both 'TMTuesday1' and 'TMTuesday2' were TMs with 20 clauses, and T and s were set to 90 and 42.3. 'TMTuesday1' produced the following results for accuracy, precision, recall and f-score : 99.84, 99.74, 97.66, 98.68. The same metrics for 'TMTuesday2' were 99.84, 99.55, 97.76, 98.64.

In comparison, when Wednesday attack data was added to the testing pool, 'TMTuesday1' and 'TMTuesday2' reported ASD of only 156.72  and 135.66 respectively. This effectively means that neither of 'TMTuesday1' or 'TMTuesday2' had learning that could effectively handle the new incoming data from Wednesday.

\subsection{Experiments with 20Newsgroup data}
\label{subsec:20NG Experiments}

\begin{table*}[ht!]
\centering
\begin{tabular}{|l|l|}
\hline
Class & Text Snippet \\
\hline
comp.graphics & \begin{tabular}[c]{@{}l@{}}``... I'm looking for shading methods and algorithms.\\ Please let me know if you know where to get source codes for that. ...''\end{tabular} \\
\hline
misc.forsale & \begin{tabular}[c]{@{}l@{}}``... I have used JVC's top of the line portable cd player for three months now.\\ I have mostly used it in my car on long trips, so it has less than 20 hours of use on it.\\ The unit is one of the best that I have seen and listened to, but\\
I am going to part with it to install a disc changer in my car. ...''\end{tabular} \\
\hline
sci.med & \begin{tabular}[c]{@{}l@{}}``...My wife cannot donate blood because she has been to a malarial region\\ 
>in the past three years.  In fact, she tried to have her bone marrow \\
>typed and they wouldn't even do that!  Why?...''\end{tabular} \\
\hline
\end{tabular}
\caption{Text snippets from examples from 3 different classes in the 20Newsgroup Dataset}
\label{tab:examplesNewsgroup}
\end{table*}

In this series of experiments, we observe how the decision making process of the TM is affected by introduction of new test data for natural language. Each data is a newsgroup post from 20 different newsgroups. In order to prevent the classifier from honing in on newsgroup names or specific email address domains to classify and instead focus on the text content itself, the headers, footers and quotes are stripped from the data using Scikit-learn. Snippets from examples of each class after stripping are given in Table \ref{tab:examplesNewsgroup}.

To begin with, we train a TM with 2 classes from the 20Newsgroup dataset : comp.graphics(Class 0) and sci.med (Class 1).
We follow a standard 80-20 split for the training and testing data. 
With number of clauses set to 500, and T and s taking values 1300 and 21.3 respectively, the TM achieves an accuracy of 83.82\%. Precision, recall and F-scores are 85.14, 84.78 and 84.77 respectively.

We try to get a sense of the decision making process on the test set, by compiling some key numbers per test example. These numbers (for each test sample) are as follows : 
\begin{itemize}
 \item Number of Clauses Triggered for Class 0 (Clause\_Cnt\_0)
    \item Number of Positive Clauses Triggered for Class 0 (Positive\_Cnt\_0)
    \item Weighted Clause Sum for Class 0 (Clause\_Sum\_0)
    \item Number of Clauses Triggered for Class 1(Clause\_Cnt\_1)
    \item Number of Positive Clauses Triggered for Class 1 (Positive\_Cnt\_1)
    \item Weighted Clause Sum for Class 1 (Clause\_Sum\_1)
    \item Absolute Difference between Clause\_Sum\_0 and \\Clause\_Sum\_1 (ASD).
\end{itemize}
For the sake of brevity, results that are tabulated in the following sections show the values only for Clause\_Cnt\_0, Positive\_Cnt\_0, Clause\_Cnt\_1, Positive\_Cnt\_1 and ASD. 

From the testing data, we obtain the average values for these parameters over all samples (Table \ref{tab:values_testing_forsale}, column `BASE'):

\setlength{\tabcolsep}{1pt}
\begin{table}[h!]
\centering
\begin{tabular}{|c|c|c|c|}
\hline
 & \textbf{\begin{tabular}[c]{@{}l@{}}`comp.graphics' \\ + `sci.med'\\ (BASE)\end{tabular}} & \textbf{\begin{tabular}[c]{@{}l@{}}Base+\\ `misc.\\ forsale'\end{tabular}} & \textbf{\begin{tabular}[c]{@{}l@{}}`misc.\\ forsale'\end{tabular}} \\ \hline
\begin{tabular}[c]{@{}l@{}}Total number of\\ test samples\end{tabular} & 785 & 1175 & 390 \\ \hline  Clause\_Cnt\_0 & 22.08 & 17.991 & 9.762 \\ \hline Positive\_Cnt\_0 & 11.777 & 10.093 & 6.703 \\ \hline Clause\_Cnt\_1 & 21.18 & 17.211 & 9.223 \\ \hline Positive\_Cnt\_1 & 10.503 & 8.387 & 4.128 \\ \hline 
ASD & 6496.162 & 5068.006 & 2193.385 \\ \hline
\end{tabular}
\caption{Values on Testing Data from different class scenarios}
\label{tab:values_testing_forsale}
\end{table}

Next, we add data from a third class to the test set and observe how the values change. The samples from this class are randomly labelled as 1 and 0. this is for the sake of the programming, because we are not concerned about the results of the classification, just the process of it. 

\subsubsection{When the Unknown data is Unrelated }
The category `misc.forsale' is definitively different from the previous two classes `comp.graphics' and `sci.med'. We choose to introduce test samples from this class as our `unseen data' to the classification model.
When the 3rd class is `misc.forsale', we observe the values as in Table \ref{tab:values_testing_forsale}, Column `Base + `misc.forsale'.

We observe a drop in the ASD values in this scenario, compared to the original two class system. This gets more interesting if we see the values for only the samples tested from the `new class' (Table \ref{tab:values_testing_forsale}, Column `misc.forsale'). The ASD for this group of samples is significantly lower than the ASD for the samples where the classes were the same as what the TM was trained on. This can be explained by the fact that when the TM tries to classify known information, the weighted sum of clauses voting for Class 0 is significantly different from the weighted sum of clauses voting for Class 1. In other words, a higher ASD indicates a more confident decision. When the TM tasked with classifying samples from a class the model has not seen before, the ASD thus becomes low, indicating that the TM did not know how to make a confident decision. We also see that, on average, lower number of clauses were involved in the decision making process in the samples from the new class compared to the 2 original classes, simply because there were less patterns overlapping between the learnt clauses and the test samples.

Moreover, the value obtained for the other calculated parameters also give insight into the learning. Clause\_Cnt\_0 and Clause\_Cnt\_1 being high indicates that there were more clauses that matched some or all of the test input and thus voted for a result. Thus when the test data is something that the model has not seen before, the number of clauses involved in the decision are lower than if the test data is known to the model. Higher ratios of Clause\_Cnt\_0 to Positive\_Cnt\_0 (and similarly the ratio of Clause\_Cnt\_1 to Positive\_Cnt\_1) signal that the model learnt more positive classifying information (i.e. it was easier to determine what patterns are positively indicative of belonging to the class). A lower ratio meant that patterns that indicated a sample to Not belong to a class were more useful in getting a prediction.

Diving a little bit deeper into this experiment, we obtain the same values, this time separated by class (both actual and predicted). They are shown in Table \ref{tab:values_testing_detailed_forsale}. What is evident here is that, the more confident the classification, the greater is the number of clauses triggered and higher the ASD, on average. Even for samples from the original 2 class set, the ones classified wrongly (false positives and false negatives), show much lower numbers of triggered clauses and ASD, than the numbers recorded for correctly classified instances.

\setlength{\tabcolsep}{1pt}
\begin{table*}[h!]
\centering
\small
\begin{tabular}{|l|c|c|c|c|c|c|}
\hline
& \textbf{\begin{tabular}[c]{@{}c@{}}Truth: 1\\ Pred: 1\end{tabular}} & \textbf{\begin{tabular}[c]{@{}c@{}}Truth: 0\\ Pred: 0\end{tabular}} & \textbf{\begin{tabular}[c]{@{}c@{}}Truth: 0\\ Pred: 1\\ (False\\Positives)\end{tabular}} & \textbf{\begin{tabular}[c]{@{}c@{}}Truth: 1\\ Pred: 0\\ (False\\negatives)\end{tabular}} & \textbf{\begin{tabular}[c]{@{}c@{}}Truth: 2\\ Pred: 1\end{tabular}} & \textbf{\begin{tabular}[c]{@{}c@{}}Truth: 2\\ Pred: 0\end{tabular}} \\ \hline
\begin{tabular}[c]{@{}l@{}}Total number of\\ test samples\end{tabular} & 351 & 328 & 61 & 45 & 151 & 239 \\ \hline Clause\_Cnt\_0 & 20.598 & 27.299 & 11.197 & 10.356 & 6.305 & 11.946 \\ \hline Positive\_Cnt\_0 & 4.231 & 21.799 & 4.852 & 6.978 & 2.947 & 9.075 \\ \hline Clause\_Cnt\_1 & 18.823 & 27.521 & 9.574 & 9.067 & 5.642 & 11.485 \\ \hline Positive\_Cnt\_1 & 16.547 & 5.704 & 6.557 & 3.689 & 4.649 & 3.799 \\ \hline ASD & 6935.815 & 7637.378 & 1439.475 & 1603.289 & 985.364 & 2956.611 \\ \hline
\end{tabular}
\caption{Detailed Values on Testing Data from different class scenarios. 0 stands for Comp.Graphics, 1 for Sci.med and 2 for Misc.forsale.}
\label{tab:values_testing_detailed_forsale}
\end{table*}

\subsubsection{When the Unknown data is Related}
Next, we repeat the same experiments as above with a set of (somewhat) related data. We choose the category `comp.os.ms-windows.misc', which is related to comp.graphics, primarily due to sharing the same super category (computers). Comparing the values obtained in this scenario versus those in the 2-class scenario gives us the Table \ref{tab:values_testing_mswindows}

\setlength{\tabcolsep}{1pt}
\begin{table}[h!]
\centering
\begin{tabular}{|c|c|c|c|}
\hline
 & \textbf{\begin{tabular}[c]{@{}l@{}}`comp.graphics' \\ + `sci.med'\\ (BASE)\end{tabular}} & \textbf{\begin{tabular}[c]{@{}l@{}}Base+\\ `comp.os.\\ms-windows'\end{tabular}} & \textbf{\begin{tabular}[c]{@{}l@{}}`comp.os.\\ms-windows'\end{tabular}} \\ \hline
\begin{tabular}[c]{@{}l@{}}Total test \\ samples\end{tabular} & 785 & 1179 & 394 \\ \hline Clause\_Cnt\_0 & 22.08 & 20.601 & 17.652 \\ \hline +ve\_Cnt\_0 & 11.777 & 11.879 & 12.081 \\ \hline Clause\_Cnt\_1 & 21.18 & 19.773 & 16.97 \\ \hline +ve\_Cnt\_1 & 10.503 & 8.859 & 5.584 \\ \hline 
ASD & 6496.162 & 5594.723 & 3798.711 \\ \hline 
\end{tabular}
\caption{Values on Testing Data from different class scenarios}
\label{tab:values_testing_mswindows}
\end{table}

Similar to the previous case, we again observe a fall in the ASD values when the previously unseen data is tested.
The detailed view (Table \ref{tab:values_testing_detailed_mswindows}), however, is more interesting. When samples of class `comp.os.ms-windows.misc' are processed by the model, on an average the ASD for predicting it as class 0 is closer to the ASD-s seen for correctly classified instances of known data. This clearly show that the model thought they were somewhat similar to Class 0 (comp.graphics), but did not have the level of support for the decision as the correctly classified samples for the known classes. the average number of clauses triggered also show the same trend. this is markedly different from the case where the unknown samples came from `misc.forsale', where the values observed for unknown data were much lower compared to that of known data, across the board.

\setlength{\tabcolsep}{1pt}
\begin{table*}[h!]
\centering
\small
\begin{tabular}{|l|c|c|c|c|c|c|}
\hline
& \textbf{\begin{tabular}[c]{@{}c@{}}Truth: 1\\ Pred: 1\end{tabular}} & \textbf{\begin{tabular}[c]{@{}c@{}}Truth: 0\\ Pred: 0\end{tabular}} & \textbf{\begin{tabular}[c]{@{}c@{}}Truth: 0\\ Pred: 1\\ (False\\Positives)\end{tabular}} & \textbf{\begin{tabular}[c]{@{}c@{}}Truth: 1\\ Pred: 0\\ (False\\negatives)\end{tabular}} & \textbf{\begin{tabular}[c]{@{}c@{}}Truth: 2\\ Pred: 1\end{tabular}} & \textbf{\begin{tabular}[c]{@{}c@{}}Truth: 2\\ Pred: 0\end{tabular}} \\ \hline
\begin{tabular}[c]{@{}l@{}}Total number of\\ test samples\end{tabular} & 351 & 328 & 61 & 45 & 102 & 292 \\ \hline Clause\_Cnt\_0 & 20.598 & 27.299 & 11.197 & 10.356 & 12.49 & 19.455 \\ \hline Positive\_Cnt\_0 & 4.231 & 21.799 & 4.852 & 6.978 & 5.196 & 14.486 \\ \hline Clause\_Cnt\_1 & 18.823 & 27.521 & 9.574 & 9.067 & 10.578 & 19.202 \\ \hline Positive\_Cnt\_1 & 16.547 & 5.704 & 6.557 & 3.689 & 6.784 & 5.164 \\ \hline 
ASD & 6935.815 & 7637.378 & 1439.475 & 1603.289 & 1616.059 & 4561.144 \\ \hline 
\end{tabular}
\caption{Detailed Values on Testing Data from different class scenarios. 0 stands for Comp.Graphics, 1 for Sci.med and 2 for comp.os.ms-windows.misc.}
\label{tab:values_testing_detailed_mswindows}
\end{table*}

\subsubsection{Analysis}
The relative differences between this simple parameters obtained from a TM classification model during testing thus proves to be a powerful indicator of the comparative similarity of the data that the TM has been trained on, and the data it is being asked to test. As a consequence, it is possible to identify whether the model is suited for the current requirements at all, and how much so. 
\subsection{Experiments with Health Data}
\label{subsec:Patient Experiments}

\begin{figure}
\centering
\includegraphics[width=.45\textwidth]{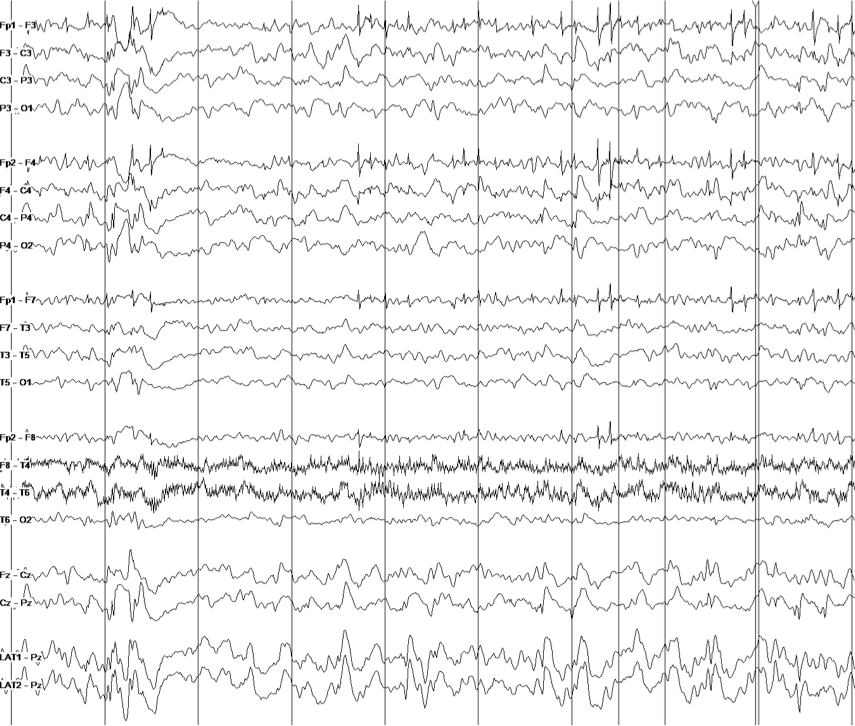} 
\caption{A sample of EEG data}
\label{fig:EEGSample}
\end{figure}

One more dataset that we try out is a patient health parameters dataset that looks to predict whether a particular patient survived or did not survive based on various collated parameters. In particular, we use electroencephalography (EEG) measures of brain activity of post-arrest patients, which and is typically monitoring across $22$ channels at $256 Hz$. A sample of EEG data is shown in Figure \ref{fig:EEGSample}. A single day’s EEG recording of one patient results in approximately $4.46 \times 10^8$ data points. Recent work has explored quantitative EEG (qEEG) characteristics and their value as predictors of eventual recovery. In this paper we will consider analysis of a single important qEEG feature: suppression ratio (SR), which quantifies the proportion of a given epoch (for example, a ten-second moving window) that is without EEG activity (i.e. suppressed).
We binarize the input variables for each patient by comparing various statistics (mean,median,variance, min,max) for the patient SR
with the corresponding values for the cohort, computed from the training set. Every variable is set to 1 if it is above the corresponding population mean, and to 0 otherwise.  

The data is heavily skewed, with the class distribution being 390 `Survived' and 935 `Did not Survive'. 
A TM equipped with 20 clauses, T= 56 and s = 11.3 reaches an accuracy of 75.57\% with precision, recall and F1-score being 69.11\%,  63.70\% and 64.96\% respectively. The class imbalance definitely contributes to such poor performance. In terms of our TM Testing parameters we obtain the values shown in Table \ref{tab:values_testing_detailed_health}. The clauses learnt are described in Table \ref{tab:clauses_patient_health_basic}. We observe a lot of repeated clauses, indicating that the TM could not find anything new to learn beyond a certain (low) number of clauses, especially for Class 1 (and Negative of Class 0).

\begin{table*}[h!]
\centering
\small
\begin{tabular}{|>{\hspace{0pt}}m{0.094\linewidth}|>{\hspace{0pt}}m{0.092\linewidth}|>{\hspace{0pt}}m{0.06\linewidth}|>{\hspace{0pt}}m{0.135\linewidth}|>{\hspace{0pt}}m{0.554\linewidth}|} 
\hline
\textbf{Class} & \textbf{P/N} & Weight & \textbf{Clauses as literal}\par{}\textbf{numbers} & \textbf{Clauses as descriptive literals} \\ 
\hline
\multirow{20}{0.094\linewidth}{\hspace{0pt}Class 0: died} & \multirow{10}{0.092\linewidth}{\hspace{0pt}Positive Clause} & W:47 & x0 & mean SR, first half \\ 
\cline{3-5}
 &  & W:21 & x0 $\wedge$  x3 & mean SR, first half $\wedge$  median SR, second half \\ 
\cline{3-5}
 &  & W:39 & x0 $\wedge$  x5 $\wedge$ ¬x1 & mean SR, first half $\wedge$  minimum SR, second half $\wedge$ ¬mean SR, second half \\ 
\cline{3-5}
 &  & W:29 & x0 $\wedge$ ¬x1 & mean SR, first half $\wedge$ ¬mean SR, second half \\ 
\cline{3-5}
 &  & W:20 & x0 $\wedge$ ¬x8 & mean SR, first half $\wedge$ ¬variance of SR, first half \\ 
\cline{3-5}
 &  & W:14 & x1 & mean SR, second half \\ 
\cline{3-5}
 &  & W:50 & x1 $\wedge$  x3 & mean SR, second half $\wedge$  median SR, second half \\ 
\cline{3-5}
 &  & W:51 & x1 $\wedge$  x3 & mean SR, second half $\wedge$  median SR, second half \\ 
\cline{3-5}
 &  & W:35 & x2 $\wedge$  x8 & median SR, first half $\wedge$  variance of SR, first half \\ 
\cline{3-5}
 &  & W:24 & x6 & maximum SR, first half \\ 
\cline{2-5}
 & \multirow{10}{0.092\linewidth}{\hspace{0pt}Negative Clause} & W:-2 &  &  \\ 
\cline{3-5}
 &  & W:-13 & x7 $\wedge$ ¬x4 $\wedge$ ¬x5 $\wedge$ ¬x8 & maximum SR, second half $\wedge$ ¬minimum SR, first half $\wedge$ ¬minimum SR, second half $\wedge$ ¬variance of SR, first half \\ 
\cline{3-5}
 &  & W:-2 & x9 & variance of SR, second half \\ 
\cline{3-5}
 &  & W:-1 & x9 & variance of SR, second half \\ 
\cline{3-5}
 &  & W:-9 & x9 & variance of SR, second half \\ 
\cline{3-5}
 &  & W:-2 & x9 & variance of SR, second half \\ 
\cline{3-5}
 &  & W:-26 & x9 $\wedge$ ¬x2 $\wedge$ ¬x5 $\wedge$ ¬x8 & variance of SR, second half $\wedge$ ¬median SR, first half $\wedge$ ¬minimum SR, second half $\wedge$ ¬variance of SR, first half \\ 
\cline{3-5}
 &  & W:-9 & x9 $\wedge$ ¬x2 $\wedge$ ¬x8 & variance of SR, second half $\wedge$ ¬median SR, first half $\wedge$ ¬variance of SR, first half \\ 
\cline{3-5}
 &  & W:-3 & ¬x2 $\wedge$ ¬x5 $\wedge$ ¬x8 & ¬median SR, first half $\wedge$ ¬minimum SR, second half $\wedge$ ¬variance of SR, first half \\ 
\cline{3-5}
 &  & W:-11 & ¬x4 & ¬minimum SR, first half \\ 
\hline
\multirow{20}{0.094\linewidth}{\hspace{0pt}Class 1: survived} & \multirow{10}{0.092\linewidth}{\hspace{0pt}Positive Clause} & W:20 & x0 $\wedge$ ¬x0 $\wedge$ ¬x2 $\wedge$ ¬x5 $\wedge$ ¬x8 & mean SR, first half $\wedge$ ¬mean SR, first half $\wedge$ ¬median SR, first half $\wedge$ ¬minimum SR, second half $\wedge$ ¬variance of SR, first half \\ 
\cline{3-5}
 &  & W:6 & x6 $\wedge$ ¬x2 & maximum SR, first half $\wedge$ ¬median SR, first half \\ 
\cline{3-5}
 &  & W:9 & x9 & variance of SR, second half \\ 
\cline{3-5}
 &  & W:4 & x9 & variance of SR, second half \\ 
\cline{3-5}
 &  & W:4 & x9 & variance of SR, second half \\ 
\cline{3-5}
 &  & W:14 & x9 $\wedge$ ¬x0 $\wedge$ ¬x5 & variance of SR, second half $\wedge$ ¬mean SR, first half $\wedge$ ¬minimum SR, second half \\ 
\cline{3-5}
 &  & W:7 & x9 $\wedge$ ¬x1 $\wedge$ ¬x3 $\wedge$ ¬x4 $\wedge$ ¬x8 & variance of SR, second half $\wedge$ ¬mean SR, second half $\wedge$ ¬median SR, second half $\wedge$ ¬minimum SR, first half $\wedge$ ¬variance of SR, first half \\ 
\cline{3-5}
 &  & W:8 & x9 $\wedge$ ¬x2 $\wedge$ ¬x8 & variance of SR, second half $\wedge$ ¬median SR, first half $\wedge$ ¬variance of SR, first half \\ 
\cline{3-5}
 &  & W:15 & x9 $\wedge$ ¬x5 $\wedge$ ¬x8 & variance of SR, second half $\wedge$ ¬minimum SR, second half $\wedge$ ¬variance of SR, first half \\ 
\cline{3-5}
 &  & W:10 & ¬x4 $\wedge$ ¬x8 & ¬minimum SR, first half $\wedge$ ¬variance of SR, first half \\ 
\cline{2-5}
 & \multirow{10}{0.092\linewidth}{\hspace{0pt}Negative Clause} & W:-36 & x0 $\wedge$  x2 $\wedge$ ¬x3 & mean SR, first half $\wedge$  median SR, first half $\wedge$ ¬median SR, second half \\ 
\cline{3-5}
 &  & W:-43 & x0 $\wedge$  x3 & mean SR, first half $\wedge$  median SR, second half \\ 
\cline{3-5}
 &  & W:-30 & x0 $\wedge$  x5 & mean SR, first half $\wedge$  minimum SR, second half \\ 
\cline{3-5}
 &  & W:-4 & x0 $\wedge$  x6 & mean SR, first half $\wedge$  maximum SR, first half \\ 
\cline{3-5}
 &  & W:-23 & x1 & mean SR, second half \\ 
\cline{3-5}
 &  & W:-33 & x1 $\wedge$  x3 & mean SR, second half $\wedge$  median SR, second half \\ 
\cline{3-5}
 &  & W:-26 & x2 $\wedge$  x5 $\wedge$ ¬x3 $\wedge$ ¬x8 & median SR, first half $\wedge$  minimum SR, second half $\wedge$ ¬median SR, second half $\wedge$ ¬variance of SR, first half \\ 
\cline{3-5}
 &  & W:-35 & x2 $\wedge$ ¬x3 & median SR, first half $\wedge$ ¬median SR, second half \\ 
\cline{3-5}
 &  & W:-9 & x2 $\wedge$ ¬x4 & median SR, first half $\wedge$ ¬minimum SR, first half \\ 
\cline{3-5}
 &  & W:-38 & x3 & median SR, second half \\
 \hline
\end{tabular}
\caption{Clauses learnt for Patient Health information}
\label{tab:clauses_patient_health_basic}
\end{table*}

\setlength{\tabcolsep}{1pt}
\begin{table}[h!]
\centering
\begin{tabular}{|l|c|c|c|c|} 
\hline
 & \begin{tabular}[c]{@{}c@{}}\textbf{Truth: 1}\\\textbf{ Pred: 1}\end{tabular} & \begin{tabular}[c]{@{}c@{}}\textbf{Truth: 0}\\\textbf{ Pred: 0}\end{tabular} & \begin{tabular}[c]{@{}c@{}}\textbf{Truth: 0}\\\textbf{ Pred: 1}\end{tabular} & \begin{tabular}[c]{@{}c@{}}\textbf{Truth: 1}\\\textbf{ Pred: 0}\end{tabular}  \\ 
\hline
\begin{tabular}[c]{@{}l@{}}Total number of\\ test samples\end{tabular} & 45 & 286 & 31 & 76 \\ \hline Clause\_Cnt\_0 & 1.0 & 3.93 & 1.0 & 0.961 \\ \hline Positive\_Cnt\_0 & 0.0 & 3.927 & 0.0 & 0.961 \\ \hline Clause\_Cnt\_1 & 3.0 & 4.402 & 3.0 & 1.053 \\ \hline Positive\_Cnt\_1 & 3.0 & 0.028 & 3.0 & 0.0 \\ \hline
ASD & 60.0 & 291.434 & 60.0 & 70.513 \\ \hline 
\end{tabular}
\caption{Detailed Values on Testing Data from imbalanced data. 0 stands for `Did Not Survive', 1 for `Survived'}
\label{tab:values_testing_detailed_health}
\end{table}

\subsubsection{Identifying Non-Conforming Information}
We use the idea of stratified k-folds to identify the data that does not conform to the model learnt by the TM classifier. Making \textit{n} random subsets of the test set and testing them, we can identify those individual elements that have the lowest ASD. This indicates that the TM consistently fails to identify them properly.

Table \ref{tab:health_various_split_onlytest} shows the results of testing when some folds of the test set are ignored. 

\setlength{\tabcolsep}{1pt}
\begin{table}[h!]
\centering
\begin{tabular}{|l|c|c|c|c|} 
\hline
\textbf{Tested On}  & \textbf{Accuracy} & \textbf{Precision} & \textbf{Recall} & \textbf{F-Score} \\ 
\hline
\textbf{Original Test Set} & 75.57 & 69.11 & 63.70 & 64.96 \\ 
\hline
\begin{tabular}[c]{@{}l@{}}\textbf{Split with Max }\\\textbf{ASD}\end{tabular} & 78.89 & 75.14 & 71.01 & 71.78 \\ 
\hline
\begin{tabular}[c]{@{}l@{}}\textbf{Splits in top 25}\\\textbf{perc. ASD}\end{tabular} & 77.98 & 73.65 & 69.96 & 70.68 \\ 
\hline
\begin{tabular}[c]{@{}l@{}}\textbf{ALL except Split }\\\textbf{with Min ASD}\end{tabular} & 77.9 & 73.55 & 69.91 & 70.63 \\ 
\hline
\begin{tabular}[c]{@{}l@{}}\textbf{ALL except Splits}\\\textbf{in bottom 25 perc}\\\textbf{ASD}\end{tabular} & 77.91 & 73.57 & 69.95 & 70.65 \\ 
\hline
\end{tabular}
\caption{Testing results on various folds of testing data with TM trained on original training data}
\label{tab:health_various_split_onlytest}
\end{table}
It would be possible to find the most common offending instances from the ignored folds that gave the best results on average over multiple runs.

\subsubsection{Mitigating Effect of Non-conforming Information}
We reduce the effect of the class imbalance, we use SMOTE \cite{chawla2002smote} to oversample the data and rebalance it, leading to an increase in performance. The results are calculated as an average over 100 different runs of oversampling, and the were as follows> accuracy: 76.55\%, precision: 73.43\%, recall: 74.09\%, f1-score: 71.86\%. The testing values for a random iteration 
are shown in Table \ref{tab:values_testing_detailed_health_oversampling}. Thus it is definitely an improvement over the original set. The same can be seen in the clauses as well, Table \ref{tab:clauses_patient_health_oversampled}. There are less repeated clauses.

\setlength{\tabcolsep}{1pt}
\begin{table}[h!]
\centering

\begin{tabular}{|l|c|c|c|c|} 
\hline
 & \begin{tabular}[c]{@{}c@{}}\textbf{Truth: 1}\\\textbf{ Pred: 1}\end{tabular} & \begin{tabular}[c]{@{}c@{}}\textbf{Truth: 0}\\\textbf{ Pred: 0}\end{tabular} & \begin{tabular}[c]{@{}c@{}}\textbf{Truth: 0}\\\textbf{ Pred: 1}\end{tabular} & \begin{tabular}[c]{@{}c@{}}\textbf{Truth: 1}\\\textbf{ Pred: 0}\end{tabular}  \\ 
\hline
\begin{tabular}[c]{@{}l@{}}Total number of\\ test samples\end{tabular} & 58 & 284 & 35 & 61 \\ \hline Clause\_Cnt\_0 & 0.776 & 3.095 & 0.857 & 0.426 \\ \hline Positive\_Cnt\_0 & 0.0 & 3.088 & 0.0 & 0.426 \\ \hline Clause\_Cnt\_1 & 0.828 & 3.732 & 0.714 & 0.541 \\ \hline Positive\_Cnt\_1 & 0.828 & 0.0 & 0.714 & 0.0 \\ \hline 
ASD & 34.655 & 129.257 & 33.286 & 18.852 \\ \hline 
\end{tabular}
\caption{Detailed Values on Testing Data from oversampled data. 0 stands for `Did Not Survive', 1 for `Survived'}
\label{tab:values_testing_detailed_health_oversampling}
\end{table}

\begin{table*}[h!]
\centering
\small
\begin{tabular}{|>{\hspace{0pt}}m{0.104\linewidth}|>{\hspace{0pt}}m{0.104\linewidth}|>{\hspace{0pt}}m{0.063\linewidth}|>{\hspace{0pt}}m{0.131\linewidth}|>{\hspace{0pt}}m{0.533\linewidth}|} 
\hline
\textbf{Class} & \textbf{P/N} & Weight & \textbf{Clauses as literal}\par{}\textbf{numbers} & \textbf{Clauses as descriptive literals} \\ 
\hline
\multirow{20}{0.104\linewidth}{\hspace{0pt}Class 0: died} & \multirow{10}{0.104\linewidth}{\hspace{0pt}Positive Clause} & W:17 & ~ x0 $\wedge$~ x6 & mean SR, first half $\wedge$~ maximum SR, first half \\ 
\cline{3-5}
 &  & W:18 & ~ x1 $\wedge$~ x5 $\wedge$~ x7 & mean SR, second half $\wedge$~ minimum SR, second half $\wedge$~ maximum SR, second half \\ 
\cline{3-5}
 &  & W:16 & ~ x1 $\wedge$~ x5 $\wedge$~ x8 & mean SR, second half $\wedge$~ minimum SR, second half $\wedge$~ variance of SR, first half \\ 
\cline{3-5}
 &  & W:29 & ~ x1 $\wedge$~ x5 $\wedge$ ¬x0 & mean SR, second half $\wedge$~ minimum SR, second half $\wedge$ ¬mean SR, first half \\ 
\cline{3-5}
 &  & W:22 & ~ x1 $\wedge$ ¬x2 & mean SR, second half $\wedge$ ¬median SR, first half \\ 
\cline{3-5}
 &  & W:25 & ~ x2 $\wedge$~ x5 $\wedge$ ¬x0 & median SR, first half $\wedge$~ minimum SR, second half $\wedge$ ¬mean SR, first half \\ 
\cline{3-5}
 &  & W:10 & ~ x3 $\wedge$~ x5 & median SR, second half $\wedge$~ minimum SR, second half \\ 
\cline{3-5}
 &  & W:18 & ~ x3 $\wedge$~ x5 & median SR, second half $\wedge$~ minimum SR, second half \\ 
\cline{3-5}
 &  & W:17 & ~ x3 $\wedge$~ x8 & median SR, second half $\wedge$~ variance of SR, first half \\ 
\cline{3-5}
 &  & W:9 & ~ x6 & maximum SR, first half \\ 
\cline{2-5}
 & \multirow{10}{0.104\linewidth}{\hspace{0pt}Negative Clause} & W:-1 & ~ x6 & maximum SR, first half \\ 
\cline{3-5}
 &  & W:-4 & ~ x9 & variance of SR, second half \\ 
\cline{3-5}
 &  & W:-9 & ~ x9 $\wedge$ ¬x0 $\wedge$ ¬x4 $\wedge$ ¬x8 & variance of SR, second half $\wedge$ ¬mean SR, first half $\wedge$ ¬minimum SR, first half $\wedge$ ¬variance of SR, first half \\ 
\cline{3-5}
 &  & W:-18 & ~ x9 $\wedge$ ¬x1 $\wedge$ ¬x2 $\wedge$ ¬x8 & variance of SR, second half $\wedge$ ¬mean SR, second half $\wedge$ ¬median SR, first half $\wedge$ ¬variance of SR, first half \\ 
\cline{3-5}
 &  & W:-9 & ~ x9 $\wedge$ ¬x2 $\wedge$ ¬x3 $\wedge$ ¬x8 & variance of SR, second half $\wedge$ ¬median SR, first half $\wedge$ ¬median SR, second half $\wedge$ ¬variance of SR, first half \\ 
\cline{3-5}
 &  & W:-13 & ~ x9 $\wedge$ ¬x2 $\wedge$ ¬x5 $\wedge$ ¬x8 & variance of SR, second half $\wedge$ ¬median SR, first half $\wedge$ ¬minimum SR, second half $\wedge$ ¬variance of SR, first half \\ 
\cline{3-5}
 &  & W:-17 & ~ x9 $\wedge$ ¬x2 $\wedge$ ¬x5 $\wedge$ ¬x8 & variance of SR, second half $\wedge$ ¬median SR, first half $\wedge$ ¬minimum SR, second half $\wedge$ ¬variance of SR, first half \\ 
\cline{3-5}
 &  & W:-17 & ~ x9 $\wedge$ ¬x8 & variance of SR, second half $\wedge$ ¬variance of SR, first half \\ 
\cline{3-5}
 &  & W:-8 & ¬x2 $\wedge$ ¬x4 $\wedge$ ¬x8 & ¬median SR, first half $\wedge$ ¬minimum SR, first half $\wedge$ ¬variance of SR, first half \\ 
\cline{3-5}
 &  & W:-18 & ¬x4 $\wedge$ ¬x8 & ¬minimum SR, first half $\wedge$ ¬variance of SR, first half \\ 
\hline
\multirow{20}{0.104\linewidth}{\hspace{0pt}Class 1: survived} & \multirow{10}{0.104\linewidth}{\hspace{0pt}Positive Clause} & W:13 & ~ x6 $\wedge$ ¬x2 $\wedge$ ¬x8 & maximum SR, first half $\wedge$ ¬median SR, first half $\wedge$ ¬variance of SR, first half \\ 
\cline{3-5}
 &  & W:15 & ~ x9 $\wedge$ ¬x2 $\wedge$ ¬x3 & variance of SR, second half $\wedge$ ¬median SR, first half $\wedge$ ¬median SR, second half \\ 
\cline{3-5}
 &  & W:14 & ~ x9 $\wedge$ ¬x2 $\wedge$ ¬x3 $\wedge$ ¬x8 & variance of SR, second half $\wedge$ ¬median SR, first half $\wedge$ ¬median SR, second half $\wedge$ ¬variance of SR, first half \\ 
\cline{3-5}
 &  & W:17 & ~ x9 $\wedge$ ¬x2 $\wedge$ ¬x8 & variance of SR, second half $\wedge$ ¬median SR, first half $\wedge$ ¬variance of SR, first half \\ 
\cline{3-5}
 &  & W:15 & ~ x9 $\wedge$ ¬x4 $\wedge$ ¬x5 & variance of SR, second half $\wedge$ ¬minimum SR, first half $\wedge$ ¬minimum SR, second half \\ 
\cline{3-5}
 &  & W:12 & ¬x0 $\wedge$ ¬x4 & ¬mean SR, first half $\wedge$ ¬minimum SR, first half \\ 
\cline{3-5}
 &  & W:22 & ¬x1 $\wedge$ ¬x2 $\wedge$ ¬x3 & ¬mean SR, second half $\wedge$ ¬median SR, first half $\wedge$ ¬median SR, second half \\ 
\cline{3-5}
 &  & W:18 & ¬x2 $\wedge$ ¬x4 $\wedge$ ¬x8 & ¬median SR, first half $\wedge$ ¬minimum SR, first half $\wedge$ ¬variance of SR, first half \\ 
\cline{3-5}
 &  & W:20 & ¬x2 $\wedge$ ¬x5 $\wedge$ ¬x8 & ¬median SR, first half $\wedge$ ¬minimum SR, second half $\wedge$ ¬variance of SR, first half \\ 
\cline{3-5}
 &  & W:25 & ¬x5 & ¬minimum SR, second half \\ 
\cline{2-5}
 & \multirow{10}{0.104\linewidth}{\hspace{0pt}Negative Clause} & W:-21 & ~ x0 $\wedge$~ x5 $\wedge$~ x8 & mean SR, first half $\wedge$~ minimum SR, second half $\wedge$~ variance of SR, first half \\ 
\cline{3-5}
 &  & W:-16 & ~ x1 & mean SR, second half \\ 
\cline{3-5}
 &  & W:-11 & ~ x1 & mean SR, second half \\ 
\cline{3-5}
 &  & W:-16 & ~ x1 $\wedge$~ x3 $\wedge$~ x5 & mean SR, second half $\wedge$~ median SR, second half $\wedge$~ minimum SR, second half \\ 
\cline{3-5}
 &  & W:-25 & ~ x1 $\wedge$~ x5 $\wedge$~ x7 & mean SR, second half $\wedge$~ minimum SR, second half $\wedge$~ maximum SR, second half \\ 
\cline{3-5}
 &  & W:-24 & ~ x1 $\wedge$~ x5 $\wedge$ ¬x4 & mean SR, second half $\wedge$~ minimum SR, second half $\wedge$ ¬minimum SR, first half \\ 
\cline{3-5}
 &  & W:-16 & ~ x1 $\wedge$~ x8 & mean SR, second half $\wedge$~ variance of SR, first half \\ 
\cline{3-5}
 &  & W:-18 & ~ x3 & median SR, second half \\ 
\cline{3-5}
 &  & W:-2 & ~ x6 & maximum SR, first half \\ 
\cline{3-5}
 &  & W:-8 & ~ x6 & maximum SR, first half \\
 \hline
 \end{tabular}
\caption{Clauses learnt for Patient Health information with Oversampling using SMOTE}
\label{tab:clauses_patient_health_oversampled}
\end{table*}

\subsubsection{Informed Oversampling}
One potential issue with oversampling is that examples that are `harder' to classify can be sampled multiple times, resulting in an increase of such problematic items. To address this, we make use of our testing values to indicate samples that are more easily classified than others.

We run a stratified k-folds selector in the input data, with 10 strata, repeated twice (therefore ending up with 20 subsets of the initial data). Performing experiments similar to those described above on these data subsets, we obtain a collection of different ASD values, each from one of the subsets. We chose the subset that produced the highest ASD to carry out oversampling and train a TM with the resultant dataset. Finally, we test on the original data with this improved TM to see if there has been any performance gains.

With the new method, the performance metrics averaged over 500 runs are : accuracy: 79.07\%, precision: 76.8\%, recall: 77.34\% and f-score: 75.24\%. The testing values of a random iteration is given in Table \ref{tab:values_testing_detailed_health_stratkmeans}.

\setlength{\tabcolsep}{1pt}
\begin{table}[h!]
\centering

\begin{tabular}{|l|c|c|c|c|} 
\hline
 & \begin{tabular}[c]{@{}c@{}}\textbf{Truth: 1}\\\textbf{ Pred: 1}\end{tabular} & \begin{tabular}[c]{@{}c@{}}\textbf{Truth: 0}\\\textbf{ Pred: 0}\end{tabular} & \begin{tabular}[c]{@{}c@{}}\textbf{Truth: 0}\\\textbf{ Pred: 1}\end{tabular} & \begin{tabular}[c]{@{}c@{}}\textbf{Truth: 1}\\\textbf{ Pred: 0}\end{tabular}  \\ 
\hline
\begin{tabular}[c]{@{}l@{}}Total number of\\ test samples\end{tabular} & 106 & 236 & 73 & 23 \\ \hline Clause\_Cnt\_0 & 2.321 & 3.669 & 2.082 & 1.304 \\ \hline Positive\_Cnt\_0 & 0.0 & 3.564 & 0.0 & 1.261 \\ \hline Clause\_Cnt\_1 & 0.358 & 5.483 & 0.315 & 1.565 \\ \hline Positive\_Cnt\_1 & 0.358 & 0.0 & 0.315 & 0.0 \\ \hline
ASD & 28.745 & 159.852 & 25.315 & 52.739 \\ \hline 
\end{tabular}
\caption{Detailed Values on Testing Data from data with informed oversampling. 0 stands for `Did Not Survive', 1 for `Survived'}
\label{tab:values_testing_detailed_health_stratkmeans}
\end{table}

Moreover, we run variants of this experiments, to explore how the classification differs. One version uses all the splits that have ASDs in the top 25th percentile to train. Another variant uses all splits to train except the one split which has the least ASD. A third variant discards those splits whose ASDs lie in the bottom 25th percentile. With each of these variants we record the results both with and without oversampling in Table \ref{tab:health_various_split_exp}.

\setlength{\tabcolsep}{1pt}
\begin{table*}[h!]
\centering
\begin{tabular}{|l|c|c|c|c|c|} 
\hline
\textbf{Trained With} & \textbf{Over-}\par{}\textbf{sampled} & \textbf{Accuracy} & \textbf{Precision} & \textbf{Recall} & \textbf{F-Score} \\ 
\hline
\textbf{Original Test Set} & NO & 75.57 & 69.11 & 63.70 & 64.96 \\ 
\hline
\textbf{Original Test Set} & YES & 76.55 & 73.43 & 74.09 & 71.86 \\ 
\hline
\begin{tabular}[c]{@{}l@{}}\textbf{Split with Max }\\\textbf{ASD}\end{tabular} & YES & 79.07 & 76.8 & 77.34 & 75.24 \\ 
\hline
\begin{tabular}[c]{@{}l@{}}\textbf{Splits in top 25}\\\textbf{perc. ASD}\end{tabular} & YES & 75.59 & 66.75 & 69.95 & 67.29 \\ 
\hline
\begin{tabular}[c]{@{}l@{}}\textbf{ALL except Split }\\\textbf{with Min ASD}\end{tabular} & YES & 76.28 & 74.54 & 79.25 & 74.35 \\ 
\hline
\begin{tabular}[c]{@{}l@{}}\textbf{ALL except Splits}\\\textbf{in bottom 25 perc}\\\textbf{ASD}\end{tabular} & YES & 75.31 & 75.01 & 73.08 & 71.95 \\ 
\hline
\begin{tabular}[c]{@{}l@{}}\textbf{Split with Max }\\\textbf{ASD}\end{tabular} & NO & 75.9 & 66 & 63.47 & 62.91 \\ 
\hline
\begin{tabular}[c]{@{}l@{}}\textbf{Splits in top 25}\\\textbf{perc. ASD}\end{tabular} & NO & 78.05 & 74.87 & 68.46 & 70.06 \\ 
\hline
\begin{tabular}[c]{@{}l@{}}\textbf{ALL except Split}\\\textbf{with Min ASD}\end{tabular} & NO & 75.19 & 66.08 & 61.52 & 60.7 \\ 
\hline
\begin{tabular}[c]{@{}l@{}}\textbf{ALL except Splits}\\\textbf{in bottom 25 perc}\\\textbf{ASD}\end{tabular} & NO & 78.05 & 73.89 & 66.07 & 67.44 \\
\hline
\end{tabular}
\caption{Testing results with TM trained on various folds of training data}
\label{tab:health_various_split_exp}
\end{table*}

\subsection{Summarizing Thoughts about Experimental Results}
We explore how the global and local explanations that a TM learns reacts to a scenario with possible dynamic data via multiple experiments. From the global perspective, if the same TM is trained side by side with data that has undergone changes, the changes in the data can be traced back from the changes observed in the learnt clauses. This allows for identification of data modifications post training, without requiring detailed prior information. In contrast, the local explanation can come useful when the training has been constant, but the samples on which prediction needs to be carried out may not be conforming to the original training data. 

Observing the decision process of the TM for individual samples indicate how well the training is equipped to handle the test. 

First, we introduce the theoretical proposal and illustrate its details with the use of a Relational Tsetlin Machine on synthetically generated Hat Passing data. A RTM arrives at a set of Horn Clauses for the initial data, which we assume to be the ground truth. Subsequently, two experiments are carried out, one of \textit{targeted destabilization} and a second of \textit{non-targeted destabilization}. In \textit{targeted destabilization}, noise is introduced in the form of data inconsistencies specifically non-conforming to a chosen Horn Clause. In this scenario, the TM was observed acknowledging the noise by lowered weights of clauses associated with the targeted Horn Clauses, or inconsistent clause structure. In \textit{non-targeted destabilization}, noise is inserted randomly in the data. Here the TM learning incorporated the noise in the form of new clauses.

This followed by experiments conducted on three varied real-world datasets :  network intrusion data, 20Newsgroup text data and patient health data. Each creates a unique scenario for the proposed method to tackle. The network intrusion data involves a lot of features, of which only few are involved in decision making. This creates challenges in comparing clauses since different scenarios make use of different subsets of literals. Hence metrics such as clause weights and average difference of class sums (ASD) are required to better track changes in learning. Of these, ASD is shown to capture differences in learning succinctly. In 20Newsgroup data, experiments were carried out by training a TM on 2 classes, and subsequently testing with 3 classes (2 classes used for training and a new class, i.e. unseen data for the classifier). Tracking ASD changes for different scenarios showed that the closer the test data is to the training data, the more similar the ASD-s are. Finally, experiments with the Patient Health data threw up a new challenging scenario : imbalanced data. Oversampling is commonly used to mitigate the negative effects of class imbalance in classification. Here, ASD was calculated on k-fold splits of the data to 'grade' the data and subsequently it was shown that oversampling from the fold(s) with high (or highest) ASD leads to better quality data than with random oversampling.

\section{Conclusion}
\label{sec:Conclusion}
In this paper we explored how the Tsetlin Machine can be applied to the task of efficient data fusion.  In particular,  we explore how the explanations in form of logical clauses that a TM learns reacts to a scenario with possible noise in dynamic data. We demonstrated that TM can recognize  the noise by lowering weights of previously learned clauses, or reflect it in the form of new clauses.

We performed a comprehensive experimental study using notably different datasets: network intrusion data, 20Newsgroup text data and patient health data. The network intrusion data involves a lot of features, of which only few are involved in decision making. This creates challenges in comparing clauses since different scenarios make use of different subsets of literals. In 20Newsgroup data, experiments were carried out by training a TM on 2 classes, and subsequently testing with 3 classes (2 classes used for training and a new class, i.e. unseen data for the classifier). Finally, experiments with the Patient Health data threw up a new challenging scenario : imbalanced data.
We demonstrated that in each of those cases our approach performs very well.

We believe that the propose approach can be efficiently used in multiple scenarios, where it is important to acknowledge, identify and mitigate the effects of data characteristics changing over time.

 \bibliographystyle{elsarticle-num} 
 \bibliography{bibliography}





\end{document}